\documentclass{article}
\pdfoutput=1

\usepackage[nonatbib, preprint]{neurips_2021}

\usepackage[utf8]{inputenc} 
\usepackage[T1]{fontenc}    
\usepackage{hyperref}       
\usepackage{url}            
\usepackage{booktabs}       
\usepackage{amsfonts}       
\usepackage{nicefrac}       
\usepackage{microtype}      

\usepackage{graphics} 				
\usepackage{animate}
\usepackage{epsfig} 				
\usepackage{epstopdf}

\usepackage{listings}
\usepackage{color}
\usepackage{nameref}
\usepackage{hyperref}
\usepackage{amsmath}	 			
\usepackage{amssymb}  				

\usepackage{dsfont}			
\usepackage{mathtools}

\usepackage{epigraph}
\usepackage{lscape}
\usepackage[]{nomencl}				
\usepackage{algorithm}
\usepackage{algorithmic}
\usepackage{multicol}
\usepackage{multirow}
\usepackage{etoolbox}

\usepackage{caption}
\usepackage{subcaption}
\usepackage{wrapfig}

\usepackage{siunitx}

\usepackage{floatflt}

\usepackage{url}

\usepackage{enumitem}

\newcommand{\kldesign}{\ensuremath{\xi}}

\newcommand{\klexpectdist}[2]{\ensuremath{\mathds{E}_{#1}\left[ #2 \right]}}
\newcommand{\klexpectdistW}[1]{\ensuremath{\mathds{E}_{#1}}}

\newcommand{\klvaluegraph}[2]{Q^{#1}_{\text{GNN}}\left( #2 \right)}
\newcommand{\klvaluegraphtarget}[2]{\hat{Q}^{#1}_{\text{GNN}}\left( #2 \right)}
\newcommand{\klpolicygraph}[2]{\pi^{#1}_{\text{GNN}}\left( #2 \right)}

\newcommand{\klpolicymlp}[2]{\pi^{#1}_{\text{MLP}}\left( #2 \right)}
\newcommand{\klvaluemlp}[2]{Q^{#1}_{\text{MLP}}\left( #2 \right)}

\newcommand{\klpolicymlpsymbol}{\pi_{\text{MLP}}}
\newcommand{\klvaluemlpsymbol}{Q_{\text{MLP}}}

\newcommand{\klvaluegraphsymbol}{Q_{\text{GNN}}}
\newcommand{\klpolicygraphsymbol}[2]{\pi_{\text{GNN}}}

\newcommand{\klgraph}{\ensuremath{\mathbf{G}}}

\newcommand{\klreplay}{\ensuremath{\mathbb{B}_{\klgraph}^{\kldesign}}}
\newcommand{\klreplaydot}{\ensuremath{\mathbb{B}_{\klgraph}^{\cdot}}}
\newcommand{\klreplayset}{\ensuremath{\widetilde{\mathbb{B}}}}

\usepackage{todonotes}

\newcommand{\etal}{et\penalty50\ al.~}
\newcommand{\iid}{i.i.d.}
\newcommand{\ie}{i.e.~}

\usepackage{colortbl}%

\newcommand{\email}[1]{\href{mailto:#1}{\nolinkurl{#1}}}

\newcommand{\link}[1]{\colora{\url{#1}}}

\newcommand{\fig}[1]{Fig.~\ref{#1}}

\title{What Robot do I Need?\\ Fast Co-Adaptation of Morphology and Control using Graph Neural Networks}

\author{%
  Kevin Sebastian Luck \\
  Institute of Perception, Action and Behaviour\\
  University of Edinburgh, United Kingdom\\
  \texttt{ksluck@ed.ac.uk} \\
  \And 
  Roberto Calandra \\
  Facebook AI Research\\
  \texttt{rcalandra@fb.com} \\
  \And
  Michael Mistry \\ 
  Institute of Perception, Action and Behaviour\\
  University of Edinburgh, United Kingdom\\
  \texttt{michael.mistry@ed.ac.uk} \\
}

\begin{document}

\maketitle


\begin{abstract}
	The co-adaptation of robot morphology and behaviour becomes increasingly important with the advent of fast 3D-manufacturing methods and efficient deep reinforcement learning algorithms. 
A major challenge for the application of co-adaptation methods to the real world is the simulation-to-reality-gap due to model and simulation inaccuracies. 
However, prior work focuses primarily on the study of evolutionary adaptation of morphologies exploiting analytical models and (differentiable) simulators with large population sizes, neglecting the existence of the simulation-to-reality-gap and the cost of manufacturing cycles in the real world. 
This paper presents a new approach combining classic high-frequency deep neural networks with computational expensive Graph Neural Networks for the data-efficient co-adaptation of agents with varying numbers of degrees-of-freedom. 
Evaluations in simulation show that the new method can co-adapt agents within such a limited number of production cycles by efficiently  combining design optimization with offline reinforcement learning, that it allows for the direct application to real-world co-adaptation tasks in future work. 
\end{abstract}


\section{Introduction}
	
	The success of deep learning enabled many advances in the field of robot learning in recent years. State-of-the-art deep reinforcement learning approaches allow us to equip robots and machines with the capability of finding novel problem solving strategies and behaviours for complex tasks in areas such as manipulation~\cite{vecerik2017leveraging, ibarz2021train}, locomotion~\cite{kohl2004policy, yang2020multi} and even flying~\cite{reddy2018glider}. 
Efforts have been undertaken to repeat this success not only in behavioural adaptation but also for the design and manufacturing processes of robots, a topic which led to the emergence of new fields of research such as evolutionary robotics \cite{bredeche2018embodied, alattas2019evolutionary}. Over time, different approaches have been developed to optimize the morphology of moving machines, some considering simple or static behaviours \cite{nygaard2021real, dinev2021co, hejna2021task} while others co-learn both the design and complex behaviour simultaneously \cite{wang2018neural, gupta2021embodied, sims1994evolving}. 
A major difficulty for this endeavor is that the processes of behaviour- and morphology-adaptation happen on different time-scales: While the adaptation of behaviour and movement strategies can be done in seconds, the change of design or morphological parameters of robots requires hours or even days for the production of a new robot body. This problem has lead to three main strategies for the adaptation of robot morphology: Algorithms optimize morphologies using (1) data generated from known analytical models or (differentiable) simulators \cite{dinev2021co, lipson2000automatic, ma2021diffaqua, spielberg2019nips, hu2019chainqueen}, (2) limited small-scale robots which can be produced within hours or minutes \cite{lipson2000automatic, hale2019robot, hale2020hardware, liao2019data}, or (3) re-configurable/ shape-shifting robots \cite{jelisavcic2017real, nygaard2021real}. 
\newline
Relying exclusively on simulations or analytical models is by far the most popular strategy, because it allows the generation of vast numbers of robot populations and generations due to today's easy access to computational resources and the possibility to mass-parallelize simulations \cite{gupta2021embodied}. However, this strategy relies on the accuracy of its models and is highly vulnerable to the so-called simulation-to-reality gap \cite{jakobi1995noise, ligot2018mimicking, rosser2020sim2real}. In fact, closing the gap between simulation and real world is an active and on-going area of research in the robot learning community \cite{ligot2020simulation,qiu2020towards,chebotar2019closing}. The sim-to-reality gap becomes even more troublesome in the case of co-adapting behaviour and morphology, since optimization and reinforcement learning algorithms alike are prone to overfit to their simulated environment \cite{rosser2020sim2real, lipson2000automatic}. 
\newline
\begin{wrapfigure}{r}{0.4\textwidth}
  \vspace{-16pt}
  \centering
    \includegraphics[width=0.39\textwidth]{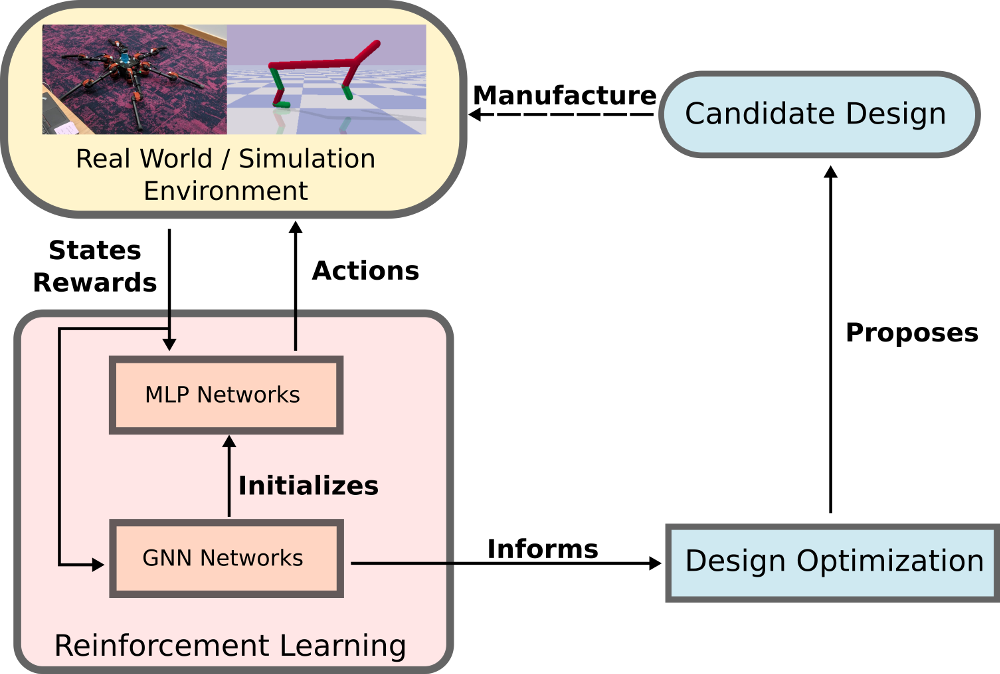}
  \caption{Proposed SG-MOrph framework in which GNNs are used to both pre-train MLP networks and inform the design optimization process. High-frequency MLP controllers are used to generate motor signals and collect experience.}
  \label{Fig::Overview}
  \vspace{-1ex}
\end{wrapfigure}%
Some work explores the possibility of applying evolutionary algorithms directly in the real world on small-scale and inexpensive robots, which can be produced quickly \cite{hale2020hardware, hale2019robot, lipson2000automatic, eiben2015evosphere, hale2019robot}. While this circumvents the sim-to-reality gap, these approaches are usually not data-efficient and rely solely on the fast-production cycles of their experimental platforms. This limits their applicability towards large-sized and expensive robots used for actual day-to-day applications in industry. This problem can be solved using robotic platforms which can reconfigure themselves, such as \cite{hornby2001evolution,nygaard2021real}. However, we argue that this places a strong restriction on the type of design optimization which are possible either by limiting choice, or by only being applicable to a selected few mechanical parameters like motor parameters or leg lengths, but not to others such as type or mixture of material~\cite{li2020review}.
Furthermore, these types of strategies often only try to improve upon the measure of computational cost or processing time (see e.g. \cite{dinev2021co}), while we find that 
\textbf{manufacturing times and cost of resources outweigh the computational cost of design optimization considerably}.
\newline
Therefore, this paper presents a new framework (Fig.\ \ref{Fig::Overview}) capable of co-adapting behaviour and design of robots with varying degrees-of-freedom (DoF) in a data-efficient way, suitable for the direct application of the method in the real world. 
The aim of the method is to be able to improve morphologies within a handful of production cycles, leveraging all data collected by combining online and offline deep reinforcement learning methods. To this end we will combine computationally expensive and slow graph neural networks (GNNs) with high-frequency neural controllers in the form of classical neural networks (MLPs). While GNNs are capable of evaluating the performance of unseen morphologies prior to production, high-frequency controllers will be used to quickly adapt the behaviour of the current prototype and avoiding problems caused by time-delays~\cite{ramstedt2019real,mahmood2018setting}.
To evaluate our approach we will study its performance in a case study in simulation, which, while not in the real world, emulates the cost of production cycles by restricting the number of possible design changes, prohibits parallel evaluations of design prototypes and considers the simulation environment as a black box.




\section{Related Work}
\label{sec:related}

	\textbf{Graph Neural Networks:}
Wang \etal \cite{wang2018nervenet} presented with Nervenet a first approach utilizing GNNs in combination with on-policy Proximal Policy Search (PPO) for learning behavioural policies and shows that their generalization capabilities are superior to MLP-based alternatives.
Huang \etal \cite{ huang2020one} extended upon this work and showed that GNNs can be used to learn policies across a large variety of morphological graph structures using a single network shared between nodes. However, real-world applicability is restricted by requiring access to the global positions of links/joints, data which would need to be acquired in the real world via motion capturing or a known dynamics model.
In a follow-up work Wang \etal \cite{wang2018neural} present a co-adaptation approach in which GNN-policies are inherited  and used as starting points for new morphologies proposed by an evolutionary algorithm manipulating the graph structures, with population sizes from 16-100. Fitness of morphologies are measured by using the episodic reward computed by simulating each design in the current population. In order to reduce the number of morphologies simulated per population by removing the worst k agents, a GNN-predictor network is used which. Given the design parameters and graph of a morphology this GNN network predicts its performance. 
Similarly, the work presented by Zhao \etal \cite{zhao2020robogrammar} learns a heuristic GNN-predictor network which estimates the episodic reward of graphs generated by a recursive graph grammar. Unlike the previously discussed approaches, which utilize PPO to learn control policies, \cite{zhao2020robogrammar} employs a model-predictive control approach using the simulator for rollouts to find optimal motor commands. 
Although not explicitly optimizing for morphologies, the approach presented in \cite{pathak2019learning} learns a GNN-policy capable of combining multiple agent bodies together into a large organism, similarly to a re-configurable robot. 
Most approaches discussed above consider either exclusively discrete search spaces for morphology optimization or utilize a combination-mutation-pertubation-like evolutionary algorithm for optimizing continuous and discrete parameters together.
\newline
\textbf{Model- or Simulation-based Adaptation:} Gupta \etal present with DERL \cite{gupta2021embodied} a co-adaptation approach relying on massive parallelization of simulated agent designs. 288 agents are evaluated and trained concurrently using a distributed version of PPO with standard MLP-policies. Morphologies are optimized using an evolutionary approach with a population size of 576 agents, which are mutated with a tournament style approach resulting in the simulation of over 4000 morphologies. 
Ha \etal present in \cite{Ha2018computational} an approach co-optimizing robot morphology and a motion plan using the implicit function theorem on the known kinematic model of the robot. Dinev \etal present a similar approach in \cite{dinev2021co} utilizing a differentiable simulator and motion planner for computing the gradient of design and control signals of a robot in order to improve the computational cost of the design optimization. Similarly, Chen \etal \cite{Cheneabg2133, chen2020hardware} present a policy-gradient-based method utilizing a differentiable hardware model as a hardware policy. 
These approaches are dependent on the accuracy of model-equations or simulators and are generally unsuitable for direct application in the real world, especially in absence of accurate analytical models and large number of manufactured designs required. 
\newline
\textbf{MLP-based approaches: }
Schaff \etal \cite{schaff2019jointly} present a policy-gradient-like strategy to optimize the parameters of a design parameter distribution from a population of design candidates which were evaluated in simulation. Similarly, Ha presents an approach \cite{ha2019reinforcement} using the REINFORCE algorithm to optimize design parameters with a population-based policy gradient method.
An approach combining a variational auto-encoder with a neural network performance predictor is presented by Pan \etal in \cite{pan2020emergent}. The method optimizes design parameters and static torque forces of a three-finger gripper grasping a variety of objects.
Chen \etal present in \cite{Cheneabg2133, chen2020hardware} a model-free variation of their approach with PPO, which uses hardware parameters as part of the action space.
Closest to our approach and a generalization of \cite{Cheneabg2133, chen2020hardware} is the algorithm presented by Luck \etal in \cite{luck2020data} which introduces the idea of using the Q-function for optimizing agent morphologies in order to decrease the number of production cycles. However, this approach uses only classical MLP neural networks and is, like all approaches discussed in this section, unable to optimize both design parameters and morphological structures, such as the number and placement of limbs or actuators. We extend the approach presented by Luck \etal \cite{luck2020data} by utilizing GNN networks for population data and a novel pre-training method for warm-starting MLP actor and critic networks on a new morphology.


\section{Background}
    \paragraph{Graph Neural Networks}
\label{sec::background::gnn}
In recent years GNNs have seen numerous applications in continuous control problems such as locomotion~\cite{wang2018nervenet, pathak2019learning} and manipulation tasks~\cite{lin2021efficient, gu2020tactilesgnet}. 
Classical neural network architectures such MLP networks have the problem that the number of input and output dimensions are fixed, i.e., one cannot easily increase or decrease the dimensionality of the task at hand during the training process, e.g. by adding or removing degrees-of-freedom (DoFs) \cite{wang2018nervenet}. 
Exploiting the fact that most simulated agents and robots have an inherent graph structure, GNNs are able to decompose the learning problem using shared neural networks and, thus, are applicable to problems where MLPs are not: Learning behavioural policies across agent morphologies with varying DOFs and state sizes. 
For the purpose of this paper we define a graph as $\mathbf{G}=(\{n_1, n_2, \cdots n_i\}, \mathbf{A})$ with nodes $n_i$ and an adjacency matrix $\mathbf{A}$. 
Without loss of generality we will throughout this paper use graphs with symmetric adjacency matrices, i.e. graphs with bi-directional edges. 
Each node $n_i$ in a graph has two fields: input data $\mathbf{s}^i$ and node-features $f^i$. 
Input data is assumed to change over time and is generally a subset of the current state $\mathbf{s}_t$ of an agent or robot.
Node features $f^i$, on the other hand, are assumed to be fixed and do not change once the graph is created. 
Similarly to prior work, we will employ three types of networks in our GNNs: processing, message-generating and output networks. 
The processing network, often also called input network, will process node data and incoming messages and computes the current state of the node $n_i$ with
$
    \mathbf{h}^i_{t+1} = \text{MLP}(\mathbf{s}^i, \mathbf{f}^i, \hat{\mathbf{m}}^i, \mathbf{h}^i_t)\,.
$
The initial node state $\mathbf{h}^i_0$ is set to zero. The message-generating network processes the node state and produces a message to be sent to all neighbours of node $n_i$. The messages are computed by
$
    \mathbf{m}^i_t = \text{MLP}(\mathbf{h}^i_t, \mathbf{e}^i)\,,
$
where $\mathbf{h}^i_t$ is the current node state and $\mathbf{e}_i$ edge-features. 
The only edge feature we will be using are directional indications of $-1$ or $1$ for edges between nodes. 
Messages $\mathbf{m}$ received by a node have to be aggregated such that they can be processed by the processing/input network. 
Several strategies exist for this, such as max-pooling and averaging. In this paper we used sum-aggregation with
$
    \hat{\mathbf{m}}_t^i = \sum_{j} \mathbf{m}^j_t\,.
$
During message passing rounds the processing networks, message generating networks and aggregation strategy are being repeatedly applied until a final latent state $\mathbf{h}^i_T$ is computed for each node.
Finally, the output of the individual nodes is generated with
$
    \mathbf{o}_i = \text{MLP}(\mathbf{h}^i_t)\,,
$
and is aggregated for the final output of the GNN using summation, averaging (in value functions) or concatenation (for policies) functions.

\paragraph{Soft Actor Critic}
Soft Actor Critic (SAC) \cite{haarnoja2018soft, haarnoja2018soft2} is a popular choice of algorithm for off-policy reinforcement learning. 
The method learns policies using stochastic neural networks, which are in effect able to adapt their exploratory noise over the course of the reinforcement learning process. 
To achieve this, SAC considers an alternative formulation of the standard RL objective and optimizes instead a maximum entropy objective given by $ \sum_{t=0}^{T} \text{r}(\mathbf{s}_t, \mathbf{a}_t) + \alpha H(\pi(\cdot\vert\mathbf{s}_t)\, $
where $\text{r}(\mathbf{s}_t, \mathbf{a}_t)$ is the reward for a given state-action combination and $H(\pi(\cdot\vert\mathbf{s}_t)$ defines the entropy of the policy, weighted by $\alpha$. 
In practice, the stochastic policy $\pi(\mathbf{a}_t\vert\mathbf{s}_t)$ is modelled with a neural network which outputs mean and standard deviations for each action dimension. 
Haarnoja \etal \cite{haarnoja2018soft} showed that this soft objective results in a new temporal-difference error for the Q-value function with
\begin{equation}
    L_{Q}(\mathbf{s}_t, \mathbf{a}_t, \mathbf{s}_{t+1}) = \frac{1}{2} \biggl\lVert \left(Q(\mathbf{s}_t, \mathbf{a}_t) - r(\mathbf{s}_t, \mathbf{a}_t) + \alpha \log \pi(\cdot \vert \mathbf{s}_{t+1}) -  \gamma \hat{Q}(\mathbf{s}_{t+1}, \hat{\mathbf{a}}) \right) \biggr\lVert^2_2\,, 
    \label{Eq::SAC::value}
\end{equation}
where $\hat{Q}$ is a target Q-function and $\hat{\mathbf{a}}\sim \pi(\cdot\vert \mathbf{s}_{t+1})$. Similarly, the loss function for training the policy network becomes
\begin{equation}
   L_{\pi}(\mathbf{s}_t) =  -Q(\mathbf{a}, \mathbf{s}_t) + \alpha \log \pi(\cdot\vert\mathbf{s}_t), \text{where~} \mathbf{a}\sim \pi(\cdot\vert\mathbf{s}_t)\,.
   \label{Eq::SAC::policy}
\end{equation}
To differentiate this objective which includes a sampled action, the reparameterization trick is utilized.

\section{Problem Statement}
  We consider an extension of the classic Markov decision process which is defined by the morphological graph structure $\klgraph$ and design parameters $\kldesign$ of an agent. 
Hence, the transition function $p(\mathbf{s}_{t+1} \vert \mathbf{s}, \mathbf{a}, \kldesign, \klgraph)$ not only depends on the current state $\mathbf{s}$ and action $\mathbf{a}$ taken, but also on the morphological parameters. 
Without loss of generality, we assume the morphological parameters of the agent to be static throughout episodes, \ie a trajectory $\mathbf{\tau}$ can be described with $(\mathbf{s}_0, \mathbf{a}_0, \mathbf{s}_1, \cdots, \mathbf{a}_{T-1}, \mathbf{s}_T, \kldesign, \klgraph)$. 
Our overarching goal is to co-adapt the behaviour and morphology of simulated and real-world robots by maximizing the objective
$  o(\pi, \kldesign, \klgraph) = \klexpectdist{\pi}{\sum_{i=0}^\infty \gamma^i r(\mathbf{s}_i, \mathbf{a}_i, \kldesign, \klgraph)
    \middle\vert \mathbf{a}_i = \pi(\mathbf{s_i})
    },$
with the discount factor $\gamma \in [0,1]$ and transitions $\mathbf{s}_{t+1} \sim p(\mathbf{s}_{t+1} \vert \mathbf{s}, \mathbf{a}, \kldesign, \klgraph)$. 
The reward function $r(\mathbf{s}_i, \mathbf{a}_i, \kldesign, \klgraph)$ might depend on the morphology $(\kldesign, \klgraph)$ of the agent, i.e. for including manufacturing costs or specific task-relevant dynamic properties of the given morphology. 
Optimizing this objective can be split into three separate problems: (1) Finding an optimal behavioural policy $\pi$, (2) finding an optimal morphological graph structure $\klgraph$ (discrete) and (3) uncovering the optimal design parameters $\kldesign$ (continuous). 
Optimizing the behaviour of an agent becomes feasible if we assume knowledge of its current morphological parameters $(\klgraph, \kldesign)$, hence the standard formulation of the policy $\pi(\mathbf{a} \vert \mathbf{s})$ becomes $\pi(\mathbf{a} \vert \mathbf{s}, \klgraph, \kldesign)$. 
Assuming access of these agent parameters for the policy, and the accompanying value function, is realistic in this context, because this set of parameters $(\kldesign, \klgraph)$ itself is required for the manufacturing process of the robot in the physical or simulated world.
However, we do not require the knowledge or description of how these morphological parameters influence the kinematics and dynamics of the agent in its environment.
This insight allows us to uncover and find a generalist policy and value function, which, if trained with enough data acquired from a variety of morphologies, is able to predict with increasing accuracy the optimal policy given an arbitrary set of morphological parameters $(\kldesign, \klgraph)$. 
Optimizing then the morphology of an agent given this generalist value function and policy becomes now a two-step process with the objective function given by
$\max_\klgraph \max_\kldesign \klexpectdist{\mathbf{s}\sim p(\mathbf{s})}{Q(\mathbf{s}, \pi(\mathbf{s}, \kldesign, \klgraph), \kldesign, \klgraph)},$
where states $\mathbf{s}$ are sampled from a fixed distribution $p(\mathbf{s})$, and in practice from a replay buffer. 
Crucially, this separates the problem of optimizing the agent morphology into a discrete optimization problem for the graph structure $\klgraph$, which requires generally evolutionary optimization strategies, and a continuous optimization problem for the design parameters $\kldesign$, which becomes differentiable when using a value function as surrogate. 
We will leverage these insights in the following section to describe a data-efficient process able to uncover optimized morphological parameters and behaviours quickly for agents with varying degrees-of-freedom.

\section{Data-Efficient Co-Adaptation with Graph Neural Networks}  
\label{sec:approach}

	This section introduces all the components of the proposed framework for fast and data-efficient co-adaptation of morphologies in a handful of production cycles.

\subsection{Graph Soft Actor Critic}
We extend SAC \cite{haarnoja2018soft} from its classic formulation for MLP networks and describe a version suitable for the application on GNNs. 
In particular, we re-formulate the original loss functions to take the individual output-generation on node-level into account. 
We assume a graph with $N$ nodes and an arbitrary number of edges between nodes, and input features $f^i$ and states/actions $\mathbf{s}^i / \mathbf{a}^i$ assigned to each node as described in section \ref{sec::background::gnn}.
The loss function for the q-value networks changes to
\begin{align}
    \nonumber
    \klexpectdistW{\klgraph, \kldesign}
    \Biggl[
        \klexpectdistW{(\mathbf{s}_t, \mathbf{a}_t, \mathbf{s}_{t+1})\in \klreplay} 
        \Biggl[
            \frac{1}{N} \sum_{i=0}^{N} \Biggl\lVert & \klvaluegraph{n_i}{\mathbf{s}_t, \mathbf{a}_t, \kldesign, \klgraph}
        + \alpha \log \klpolicygraph{n_i}{\mathbf{a}_{t+1} \vert \mathbf{s}_{t+1}, \kldesign, \klgraph} 
         \\
       & - \frac{1}{N} \sum_{j=0}^{N} \klvaluegraphtarget{n_j}{\mathbf{s}_{t+1}, \mathbf{a}_{t+1}, \kldesign, \klgraph} \Biggl\rVert_2^2 
       \Biggr] \Biggr]\,,
      \label{Eq::GNN::value}
\end{align}
where $\mathbf{a}_{t+1} = \klpolicygraph{}{\mathbf{a} \vert \mathbf{s}_{t+1}, \kldesign, \klgraph}$ holds, and $\klvaluegraphtarget{n_j}{\mathbf{s}_{t+1}, \mathbf{a}_{t+1}, \kldesign, \klgraph}$ is a target network.
Similarly, the resulting policy loss $L_\pi$ becomes
\begin{align}
    \label{Eq::GNN::policy}
    &\klexpectdistW{\klgraph, \kldesign}
    \left[ 
      \klexpectdistW{(\mathbf{s}_t, \mathbf{a}_t, \mathbf{s}_{t+1})\in \klreplay}
      \left[ 
        - \frac{1}{N} \sum^N_{i=0} 
          \klvaluegraph{n_i}{\mathbf{s}_{t}, \klpolicygraph{}{\mathbf{a} \vert \mathbf{s}_t, \kldesign, \klgraph}, \kldesign, \klgraph}
        - \alpha \log \klpolicygraph{n_i}{\mathbf{a} \vert \mathbf{s}_t, \kldesign, \klgraph}
      \right]
    \right] 
\end{align}
where actions are sampled from the policy distributions. 
For optimal training progress stochastic gradients are computed by averaging individual losses over multiple batches from different morphological parameters $(\kldesign, \klgraph)$. 
In our experiments we will average gradients over transition-batches of size $128$ from four randomly selected morphologies $(\kldesign, \klgraph)$, in addition to data generated from the current morphology. 

\subsection{Generalist and Specialist networks}

Following the general idea presented in \cite{luck2020data} we will use two sets of network: Generalists networks are learning policy and value-functions applicable to any possible morphology $(\kldesign, \klgraph)$ in a zero-shot manner using offline reinforcement learning. 
Specialist networks, on the other hand, are policy and value networks which will exclusively train from data acquired from the current morphology or robot in simulation/real-world. 
The presented approach will utilize classic feedforward neural networks as specialist networks and GNNs for the generalist networks. 
Using two vastly different architectures, here MLPs and GNNs, prohibits us from initializing the specialist networks with the weights from the generalists networks, a strategy used in prior work. 
However, using MLP networks enables us to circumvent two general drawbacks of GNN networks: Higher inference time and slower training processes. 
In summary, this approach exploits the idea of having fast specialist networks which can adapt quickly to the current task and morphology at hand, while the slower GNN networks act as a generalist, trying to uncover universal policies and value functions applicable across a variety of different morphologies in an offline RL setup. 
To enable the transfer of knowledge and avoid training controllers on synthesized hardware from scratch, we introduce in the following section a transfer-learning approach which will pre-train specialist networks with data generated from generalists networks. 

\subsection{GNN-to-MLP Transfer Learning for unseen Morphologies}
While GNNs are able to generalize across agent morphologies with varying DoFs, this ability comes at a cost. 
Due to their complex architecture and the required message-passing-rounds, GNNs have considerably larger computational costs regarding inference and training compared to simple MLP-networks. 
Hence, we propose to distill and transfer the knowledge encoded in the policy and value-function networks of a GNN to a classic MLP neural network. 
Given a new morphology, we will query the GNN network for its recommendation of actions and values at given states, and pre-train the MLP network with these estimates. 
These states can be generated or \textit{imagined} in two ways: (1) States can be sampled from a known state distribution $\mathbf{s}\sim p(\mathbf{s}\vert \kldesign, \klgraph)$, or (2) states and sensory information can be reused from previously seen morphologies with a similar graph structure $\klgraph$, \ie number of DoFs, but different designs $\kldesign$. For this paper we will use the latter strategy. 
For utilizing GNN-data and training the MLP network we introduce two loss functions for policy and Q-value networks. 
Due to the stochastic nature of the policy we utilize a Kullback-Leibler divergence loss for matching the moments of the Gaussian policy distributions with
\begin{equation}
    L^\pi_\text{Transf}(\mathbf{s}) = \text{KL}\left(\klpolicymlp{}{\cdot \vert \mathbf{s}} \parallel \klpolicygraph{}{\cdot \vert \mathbf{s}, \kldesign, \klgraph} \right),
    \label{Eq::Transf::policy}
\end{equation}
where the state $\mathbf{s}$ is randomly sampled from replay buffers $\klreplaydot$, i.e. replay buffers which contain data acquired from previously encountered morphologies with a similar graph structure $\klgraph$. The loss is minimized by updating and differentiating with respect to the parameters of the policy $\klpolicymlp{}{\cdot \vert \mathbf{s}}$. 
Similarly, an L2-loss is deployed for matching the Q-values of the GNN with
\begin{equation}
    L^Q_\text{Transf}(\mathbf{s}, \mathbf{a}) = \Biggl\Vert \frac{1}{N}\sum_{i=0}^{N} \klvaluegraph{n_i}{\mathbf{s}, \mathbf{a}, \kldesign, \klgraph} - \klvaluemlp{}{\mathbf{s}, \mathbf{a}} \Biggr\Vert^2_2,
    \label{Eq::Transf::value}
\end{equation}
where we compute the expected Q-value from the GNN by averaging the node-level Q-values and differentiate the loss with respect to the parameters of $\klvaluemlp{}{\mathbf{s}, \mathbf{a}}$.
Furthermore, we note that in the special case of SAC we use the weight $\alpha_\text{ML} = \frac{1}{N} \alpha_\text{GNN}$ for the MLP version of SAC, to take into account that the individual log-probabilities of each action $\mathbf{a} \sim \klpolicymlp{}{\cdot \vert \mathbf{s}}$ are summed over the action-dimensions. 
This matches the weight, i.e. temperature, of the expected entropy term between the MLP and GNN version of SAC, and prevents instabilities in the learning process. 

\subsection{Morphology Optimization}
The problem of optimizing the morphological parameters $(\klgraph, \kldesign)$ of an agent can be decomposed into a mixed discrete-continuous optimization problem. 
Our framework assumes that we have a pool of candidate graphs $P_\klgraph = (\klgraph_0, \klgraph_1, \cdots, \klgraph_k)$ which describe the number, type and connection between limbs or body-parts of the agent. 
This pool can be fixed, i.e. we are interested in selecting an optimal morphology from a restricted set, or new graphs can be added, removed or changed over time. 
To impose a ranking based on the individual performances in an environment or task, we need to furthermore find the optimal set of continuous design parameters $\kldesign_i$ for each graph $\klgraph_i$. 
Using the generalists networks learned with Graph-SAC, we get access to the following objective function
\begin{equation}
    o(\kldesign, \klgraph) = \klexpectdist{\mathbf{s} \in \klreplaydot}{
    \frac{1}{N} \sum_{i=0}^N \klvaluegraph{n_i}{\mathbf{s}, 
      \mathbf{a}, \kldesign, \klgraph
    }
    },~\text{where~} \mathbf{a}_i = \klpolicygraph{n_i}{\mathbf{s}, \kldesign, \klgraph},
    \label{Eq::mopt}
\end{equation}
with $N$ being the number of nodes in graph $\klgraph$. 
Using this objective function and any continuous optimization algorithm of our choice we are now able to find the best design parameters $\kldesign_i$ for each graph $\klgraph_i$ and order our candidate population based on the estimated performance with $\hat{P}_\klgraph = \left( (\klgraph_0, \kldesign_0, o(\klgraph_0, \kldesign_0)), (\klgraph_1, \kldesign_1, o(\klgraph_1, \kldesign_1)), \cdots\right)$ for which holds that $o(\klgraph_i, \kldesign_i) \geq o(\klgraph_{i+1}, \kldesign_{i+1})$. 
Furthermore, the objective function will be computed over a batch of randomly selected states from replay buffers $\klreplaydot$ matching the graph $\klgraph$. 
For an unbiased objective we use random states from the initial state distribution $\mathbf{s}\sim p(\mathbf{s}\vert\kldesign,\klgraph)$ which are \iid~among morphologies. 
Using states randomly selected in time from prior experience risks biasing the objective function towards the morphologies from which these states were selected from. 
Using this ranking we can now select the morphological set of parameters $(\klgraph, \kldesign)$ to be simulated or manufactured in the real world. 
This selection can either be greedy, by selecting always the highest ranked set, or non-greedy by using a selection strategy which assigns a selection-probability to each set of parameters based on its ranking. 
In our experiments we deploy a selection strategy which assigns a selection-probability to each morphology based on its rank $r$ and is proportional to $f(r) = \log(k + 2) - \log(r)$, with $k$ being the total number of available morphologies.
\subsection{Fast Co-Adaptation with GNNs}
\begin{wrapfigure}{r}{0.415\textwidth}
\begin{minipage}{0.415\textwidth}
\vspace{-5ex}
\begin{algorithm}[H]
\algsetup{linenosize=\scriptsize}
\scriptsize
\begin{algorithmic}[1]
    \STATE Initialize empty set of replay buffers $\klreplayset$.
    \STATE Select initial morphology $(\kldesign, \klgraph)$
    \FOR{$i \in (1, 2, \cdots, M)$}
        \STATE Initialize replay buffer $\klreplay$ and add replay to set $\klreplayset$
        \STATE Initialize and transfer policy and q-values from generalists networks $\klpolicygraphsymbol{}{}, \klvaluegraphsymbol$ to specialists networks $\klpolicymlpsymbol, \klvaluemlpsymbol$ with Eq. ( \ref{Eq::Transf::policy} - \ref{Eq::Transf::value})
        
        \FOR{$x$ episodes}
            \STATE Collect training experience $\mathbf{\tau} = (\mathbf{s}_0,\allowbreak \mathbf{a}_0,\allowbreak \mathbf{r}_0,\allowbreak \mathbf{s}_1,\allowbreak \cdots,\allowbreak \mathbf{s}_T,\allowbreak \kldesign,\allowbreak \klgraph)$ by executing one episode in simulated/real world environments
            \STATE Add $(\mathbf{s}_i, \mathbf{a}_i, \mathbf{r}_i, \mathbf{s}_{i+1})$ to replay buffer $\klreplay$.
            \STATE Train specialists networks $\klpolicymlpsymbol$ and $\klvaluemlpsymbol$ with random batches from replay buffer $\klreplay$ and loss functions (\ref{Eq::SAC::value} - $\ref{Eq::SAC::policy}$)
            \STATE Train generalists networks with random batches from $l$ random replay buffers sampled from the replay set $\klreplayset$ and loss functions (\ref{Eq::GNN::value} - \ref{Eq::GNN::policy})
        \ENDFOR
        \IF{Morphology Exploitation}
            \STATE Initialize empty ranking list $\widetilde{P}_\klgraph$
            \FOR{$\klgraph_i \in P_\klgraph $}
              \STATE Find design parameters $\kldesign_i$ by optimizing $\arg\max_{\kldesign_i} o(\kldesign_i, \klgraph_i)$ from Eq. (\ref{Eq::mopt}) with an continuous optimization algorithm
              \STATE Add $(\klgraph_i, \kldesign_i, o(\kldesign_i, \klgraph_i)$ to $\widetilde{P}_\klgraph$
            \ENDFOR
            \STATE Select new morphology $(\kldesign, \klgraph)$ from ranked list $\widetilde{P}_\klgraph$ with preferred selection mechanism
            \STATE \textit{Optional: } Generate new graph population $P_\klgraph$ with evolutionary or discrete optimization algorithm using ranking and estimated performances from $\widetilde{P}_\klgraph$
        \ELSE
            \STATE \textit{Morphology Exploration}: Sample design $\kldesign$ and $\klgraph$ randomly
        \ENDIF
    \ENDFOR
\end{algorithmic}
\caption{SG-MOrph}
\label{Alg::algorithm}
\end{algorithm}
\vspace{-8ex}
\end{minipage}
\end{wrapfigure}

Combining the techniques described in the previous sections gives rise to the proposed \textit{\textbf{S}pecialist-\textbf{G}eneralist \textbf{M}orphology \textbf{O}ptimization} (SG-MOrph) framework which is presented in Algorithm 1. 
Overall, the algorithm deploys two sets of networks, specialist MLP networks which are exlusively 
trained on data generated from the current synthesized morphology, and generalist GNN networks which learn universal policy and value functions applicable to a variety of morphologies. 
Specialist MLP networks are pre-trained on their designated morphology by generating \textit{imagined} q-values and policy-actions to avoid training from scratch for new morphologies. 
One of the main drivers of data-efficiency is the idea to store all data generated from observed morphologies in an array of replay buffers $\klreplayset$. 
This enables us to train the generalists networks continuously in an off-policy fashion and learn a value-function with deep reinforcement learning capable of predicting performances of unseen morphologies.


\section{Experiments}
\label{sec:result}

\subsection{Experimental Setup} 
\paragraph{Simulation Setup:}
For practical and logistical reasons evaluations were performed in simulation. 
However, the setup and access to simulated environments were artificially restricted to model the challenges one would encounter with real world environments. 
Here, hardware resources are limited and producing new robot designs is a time-intensive endeavor, prohibiting rapid switches between designs.  
For the simulation of agent designs we used the PyBullet physics engine~\cite{coumans2020} due to its open-source license and development process. 

\paragraph{Environments:}

All environments contain a number of different morphologies with varying DoFs as graph structures, which can be further individualized by selecting continuous design parameters. Thus, while the number of graph structures is limited, the number of morphology/design combinations is infinite.
Episodes in each environment are 600 steps long, and step-wise rewards are provided.
\newline
\textit{Half-Cheetah: } In our first environment we aim to co-adapt the behaviour and morphology of Half-Cheetah-inspired agents, as used in the original PyBullet environment. We considered seven different morphologies ranging from two to six DoF's with varying numbers of front and back limb-segments. The design parameters optimized were the length of each actuated limb-segment in the range of $[0.2, 0.6]$. The reward function is given by $r(\mathbf{s}) = \max(\frac{\Delta x}{10},0)$. The episode ends early with a reward of $-1$ if the absolute pitch of the torso exceeds $1.25$ radians.
\newline
\textit{Crawler: } The Crawler environment contains a mix of five different uni- and bi-pedal agents, which in practice lead to crawling behaviours. Here we optimized both limb-lengths and -orientations, with lengths in the range of $[0.2, 0.6]$ and limb-orientations in the range of $[-1.0, 1.0]$ radians. The reward function is the same as above.
\newline
\textit{Multi-Ped: } In the Multi-Ped environment we had a mix of different bi-, tri- and quad-pedal agent morphologies, for which we optimized limb lengths and orientations similarly to Crawler. %
\newline %
\begin{wrapfigure}{r}{0.37\textwidth}
        \centering
        \vspace{-10pt}
        \includegraphics[width=0.37\textwidth]{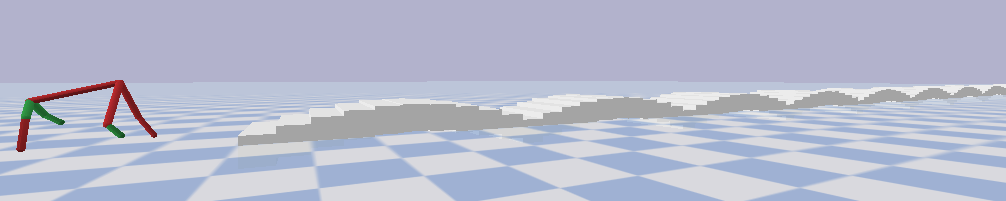}
        \caption{Simulation environment for Multi-Ped+Stairs.}
        \label{Fig::Stairs}
        \vspace{-2ex}
\end{wrapfigure}%
\textit{Multi-Ped+Stairs: } This environment (\fig{Fig::Stairs}) uses the same graph morphologies as Multi-Ped, but instead of learning locomotion strategies on flat terrain, the agent has to navigate a stairs-like pattern of blocks. The state space does not contain any information about the current positions or heights of blocks making the agent navigate the terrain blindly (similarly to \cite{siekmann2021blind}). The rewards and design parameters are the same as the Multi-Ped environment. 

\paragraph{Comparisons and Ablation Study:}
\begin{figure}[t]
    \centering
    \begin{subfigure}[t]{0.43\textwidth}
        \includegraphics[width=\textwidth]{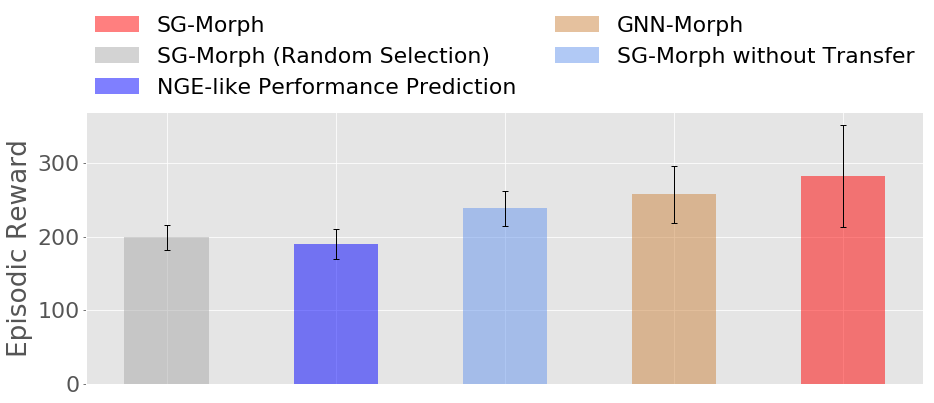}
        \caption{Half-Cheetah}
    \end{subfigure}%
    ~
    \begin{subfigure}[t]{0.175\textwidth}
        \includegraphics[height=2.6cm]{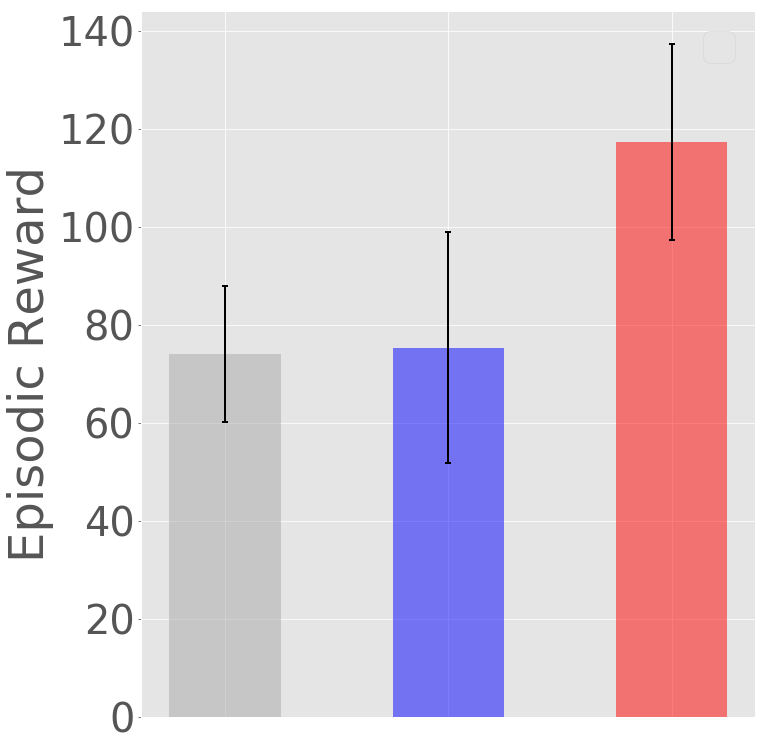}
        \caption{Crawler}
    \end{subfigure}%
    ~
    \begin{subfigure}[t]{0.175\textwidth}
        \centering
        \includegraphics[height=2.6cm]{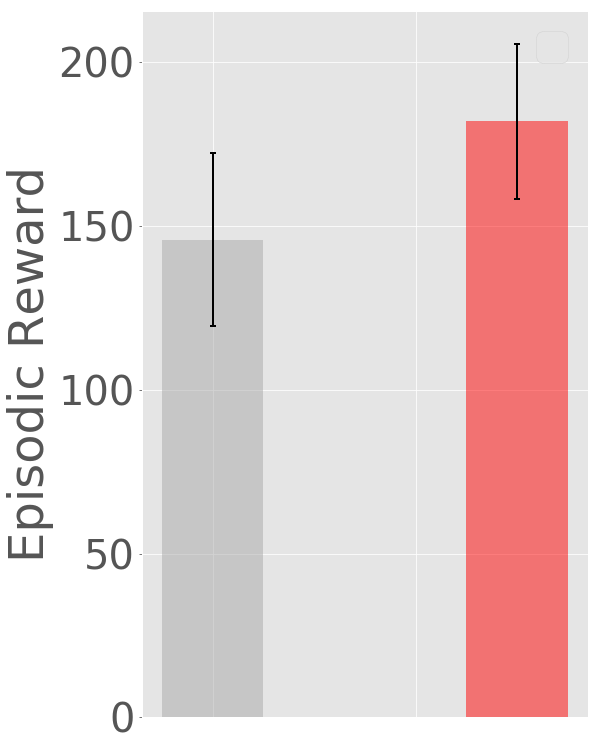}
        \caption{Multi-Ped}
    \end{subfigure}%
    ~
    \begin{subfigure}[t]{0.185\textwidth}
        \centering
        \includegraphics[height=2.6cm]{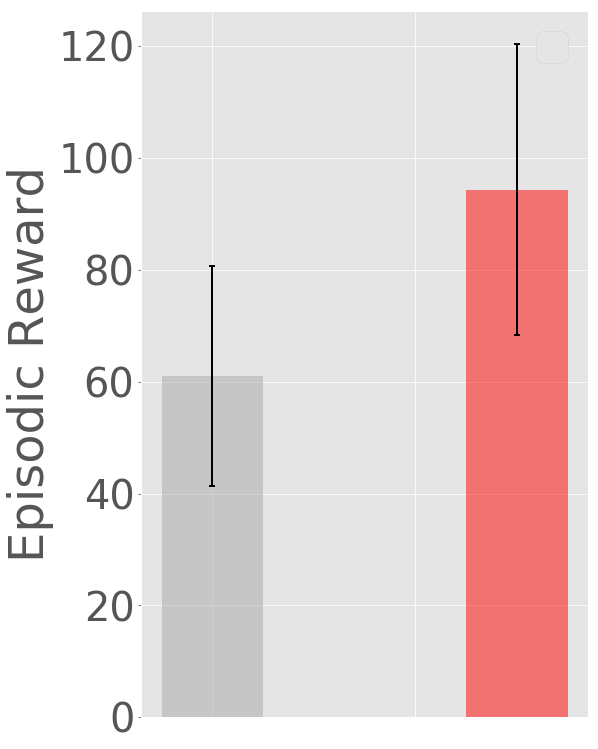}
        \caption{Multi-Ped+Stairs}
    \end{subfigure}%
    \caption{Average performance and standard deviation of the top three designs found per experiment by each method after 25 design optimization iterations and 25 randomly selected designs. Ten experiments were performed for each method in the Half-Cheetah environment and five in other environments.}
    \label{Fig:Result}
    \vspace{-3.5ex}
\end{figure} %
\begin{wrapfigure}{r}{0.32\textwidth}
    \vspace{-2ex}
    \centering
    \includegraphics[width=0.32\textwidth]{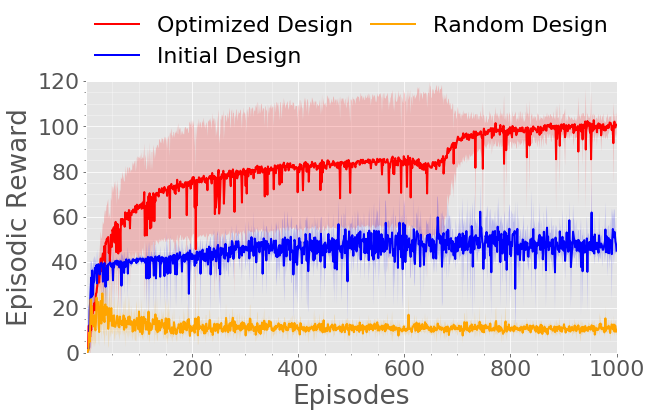}
    \caption{Performances of five MLP policies learned for an optimized design (\fig{Fig:Optimized}-d), best initial design and random design with the same number of DoF on the Multi-Ped+Stairs task. Policies were trained from scratch with SAC for 1000 episodes.}
    \label{Fig:StairsPerf}
    \vspace{-2ex}
\end{wrapfigure}%
For a better and fair comparison between all algorithms we selected two sets of design parameters from a uniform distribution for each morphology. For these fixed initial morphologies we trained standard SAC with MLP networks for 50 episodes before starting with the respective morphology optimization process. 
Thereafter, we performed 25 design optimization cycles and 25 random design selections in alternating fashion for each algorithm. Once a design is selected, 100 episodes are executed. At the end of each episode we train all MLP networks for 1,000 iterations and all GNN networks for 100 iterations. 
Experiments were conducted on an Ubuntu machine with \textit{Intel Xeon Gold 6252 (2.1GHz)} CPU and an \textit{Nvidia RTX 2080 Ti} GPU. The training time for MLP network training was $11$ seconds and for GNN networks $12$ seconds per episode. Design optimization requires an estimated $5$ minutes. 
The total runtime of each experiment was $67$ hours.
\newline
\textit{SG-MOrph:}
SG-MOrph as presented in Algorithm 1 was implemented using the Coadaptation suite~\cite{coadaptation} and the SAC implementation in PyTorch~\cite{paszke2019pytorch} provided by rlkit~\cite{rlkit}, which was also adapted for the extension of SAC with Graph Neural Networks. For training GNN networks the Deep Graph Library~\cite{wang2019deep} was utilized. 
\newline 
\textit{GNN-Predictor:}
Prior work \cite{wang2018neural, zhao2020robogrammar} used prediction networks which, given a design/morphology $(\klgraph, \kldesign)$, would estimate their fitness value. This differs to our proposed approach, which utilizes states and design/morphological parameters $(\mathbf{s}, \klgraph, \kldesign)$ as input. 
For this baseline we trained a GNN-prediction network in SG-MOrph as well, and used it during the design optimization phase. In order to train this network a dataset is collected during training which contains the graph, design parameters and the maximal achieved performance of a morphology. 
\newline
\textit{GNN-only:}
To investigate whether using a specialist MLP network trained exclusively on the current morphology improves performance, we set up a baseline similar to the approach presented in \cite{huang2020one} by training and collecting data only with a single GNN-SAC network.
\newline
\textit{SG-MOrph without Transfer:}
This variation of SG-MOrph does not initialize the MLP networks with data generated by the GNN generalists networks. Training the MLP-actor and -critic network will start from scratch for each new morphology.
\newline
\textit{Random Morphology Selection}
To test whether the design optimization truly finds improved designs we use a random baseline which uniformly selects morphology/design combinations for each environment. To give this baseline the best possible outcome we will evaluate it in parallel to the proposed SG-MOrph algorithm, i.e. after each optimized morphology we will select one randomly. This means that the random baseline should be able to benefit from any effect or insight from learning and collecting data from optimized designs.

\subsection{Results}

\begin{wrapfigure}{r}{0.4\textwidth}
    \centering
    \begin{subfigure}[t]{0.18\textwidth}
        \centering
        \includegraphics[width=\textwidth]{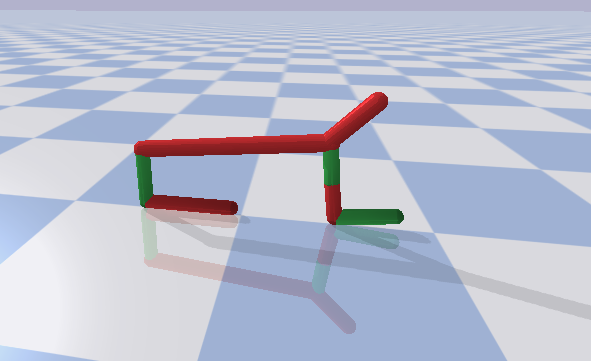}
        \caption{Half-Cheetah}
    \end{subfigure}%
    ~ 
    \begin{subfigure}[t]{0.18\textwidth}
        \centering
        \includegraphics[width=\textwidth]{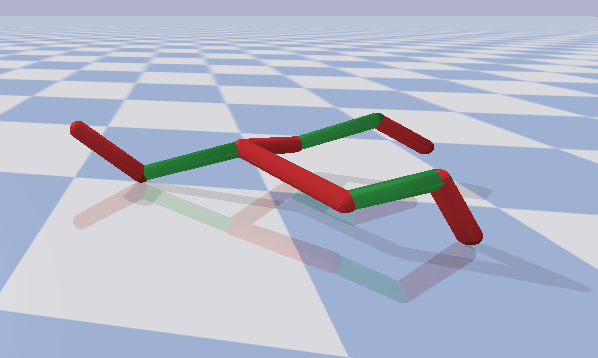}
        \caption{Crawler}
    \end{subfigure}
    
    \begin{subfigure}[t]{0.18\textwidth}
        \centering
        \includegraphics[width=\textwidth]{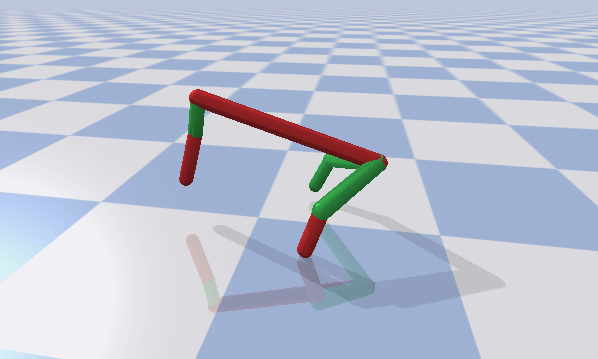}
        \caption{Multi-Ped}
    \end{subfigure}%
    ~
    \begin{subfigure}[t]{0.19\textwidth}
        \centering
        \includegraphics[width=\textwidth]{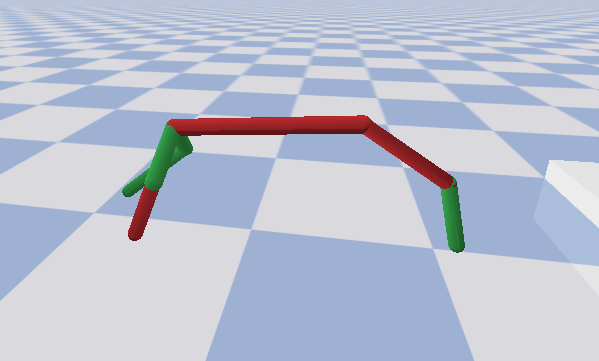}
        \caption{Multi-Ped+Stairs}
    \end{subfigure}%
    \caption{Selection of optimized  morphologies found by SG-MOrph.}
    \label{Fig:Optimized}
    \vspace{-2ex}
\end{wrapfigure}

Across all environments we found SG-MOrph to be able to uncover improved morphologies (\fig{Fig:Optimized}) and behavioural policies with better performance than the baseline methods (\fig{Fig:Result}). 
As expected, the performance difference between randomly selecting morphologies and optimizing them with SG-MOrph is the most significant. 

We find a slight improvement in using MLPs for fast adaptation on current morphologies over a GNN-only approach in 
the Half-Cheetah task (Fig.\ \ref{Fig:Result}-a), indicating a further benefit of this strategy in addition to faster training and inference times of MLP-policies. 
Results on Half-Cheetah and Crawler also indicate that the proposed use of a GNN-Value-function for estimating fitness of 
morphologies outperforms the standard approach of using a GNN-prediction network, such as used in \cite{wang2018neural} (\fig{Fig:Result}-a and \fig{Fig:Result}-b). 
One of the main reasons for this is the much larger dataset available when training a Value function over all morphologies: 
While standard reinforcement learning is able to leverage $60,000$ samples per morphology, a prediction network which fits graph and design parameters to performance only receives one data point, the best performance, per morphology. 
This causes generalization and overfitting problems in a training regime which aims to minimize the number of manufacturing cycles and assumes no availability of prior data. 
A comparison between the performance of an optimal design (\fig{Fig:Optimized}-d) found by SG-Morph 
for the Multi-Ped+Stairs task (\fig{Fig::Stairs}), the best initial design, and a randomly selected design for a morphology with five DoFs can be found in \fig{Fig:StairsPerf}. 
In this evaluation, MLP policies were trained with SAC from scratch for 1000 episodes. The mean and standard deviation of the episodic reward for each design show that the morphology-design combination uncovered by SG-MOrph performs significantly better than the initial and randomly selected designs. 
Further analysis of the data shows that the specialist MLP networks show generally better performance than the generalist GNN policies trained in SG-MOrph, with the exception of the Multi-Ped+Stairs task (see Appendix). This is a strong indication that the information transfer between MLP and GNN networks, which is in the current verison of SG-MOrph restricted to the initialization period of the MLP networks, would benefit from a continuous flow of information from the GNN to MLP networks throughout the full training process. 


\section{Conclusion}
\label{sec:conclusion}

	We introduced a new co-adaptation algorithm named SG-Morph which aims to find optimal morphologies and neural controllers while minimizing the number of morphologies necessary to be simulated or manufactured in the real world. 
This is achieved by a closed-loop between deep reinforcement learning algorithms and morphology optimization in which the value function learned from experience is leveraged to compute estimated performances of agents without the need to simulate or construct them prior to deciding on an optimal design. 
Using Graph Neural Networks and an effective transfer-learning approach for MLP-controllers allows for the optimization of morphologies with varying DoFs as well as different placements of limbs. 
Real-World applicability is ensured by not leveraging the ability to mass-parallelize simulators, thus enabling us to apply this algorithm directly to the real world in future experiments and circumvent the simulation-to-reality-gap. 
Furthermore, we use high-frequency MLP-based policies for collecting experience and learning behaviours on agents to mitigate the issue of slower inference and higher computational cost of GNNs, which would otherwise cause significant issues on real world locomotion tasks.  
Experiments show that the proposed algorithm is able to uncover better performing morphologies in a handful of production cycles and outperforms previous approaches using GNN-networks predicting performance from morphological parameters alone. 
Future work will be the application of the proposed method on an adaptable robot in the real world leveraging only real world data. This was not possible at this time due to the ongoing pandemic situation. 
A current shortfall and potential approach for further improvement is to ensure continuous flow of information from the generalist GNN-networks to the specialist MLP-networks for a new morphology in addition to the presented transfer-learning approach at the very start of the training process. 
While we hope that the proposed framework leads to new application areas for design optimization in real robots, the obvious danger of automation bias \cite{mosier2017automation} exists insofar that only designs proposed by the algorithm could be produced by operators, neglecting potential insights or skills of human designers. Hence, we believe that to achieve the full potential  of the proposed method, a close human-machine integration is necessary. 


\begin{ack}
Kevin Sebastian Luck and Michael Mistry were supported by the EPSRC UK RAI Hub NCNR (EPR02572X/1). Michael Mistry was also supported by the THING project in EU Horizon 2020 (ICT-2017-1).
The authors would like to thank Dr.\ Tamara Jacqueline Luck for comments that helped to improve the manuscript.
\end{ack}


\bibliographystyle{abbrv}
\bibliography{ms}

\clearpage
\section{Appendix}

\subsection{Morphology vs.\ Design}
This paper considers the problem of finding the optimal combination of continuous design parameters and morphological structure. 
Due to the lack of a better descriptive term, we will generally refer to \textit{design} as a set of fixed design parameters $\kldesign\in\mathbb{R}^d$ for a specific morphology $\klgraph$. 
On the other hand, we use the term \textit{morphology} or \textit{morphological structure} to refer to the number of and connections between actuators/motors in an agent. 
This means, that a graph $\klgraph$, void of any additional data and consisting only of nodes and edges, describes the morphology of an agent by stating how many actuators (\ie nodes) the agent has and how the actuators are connected to each other (\ie edges). 
Properties of motors or properties of the connecting elements (\ie limb-segments) are exclusively described by the continuous design parameters $\kldesign$. 

\subsection{SG-MOrph Details}
\begin{figure*}[h!]
    \centering
    \includegraphics[width=0.98\textwidth]{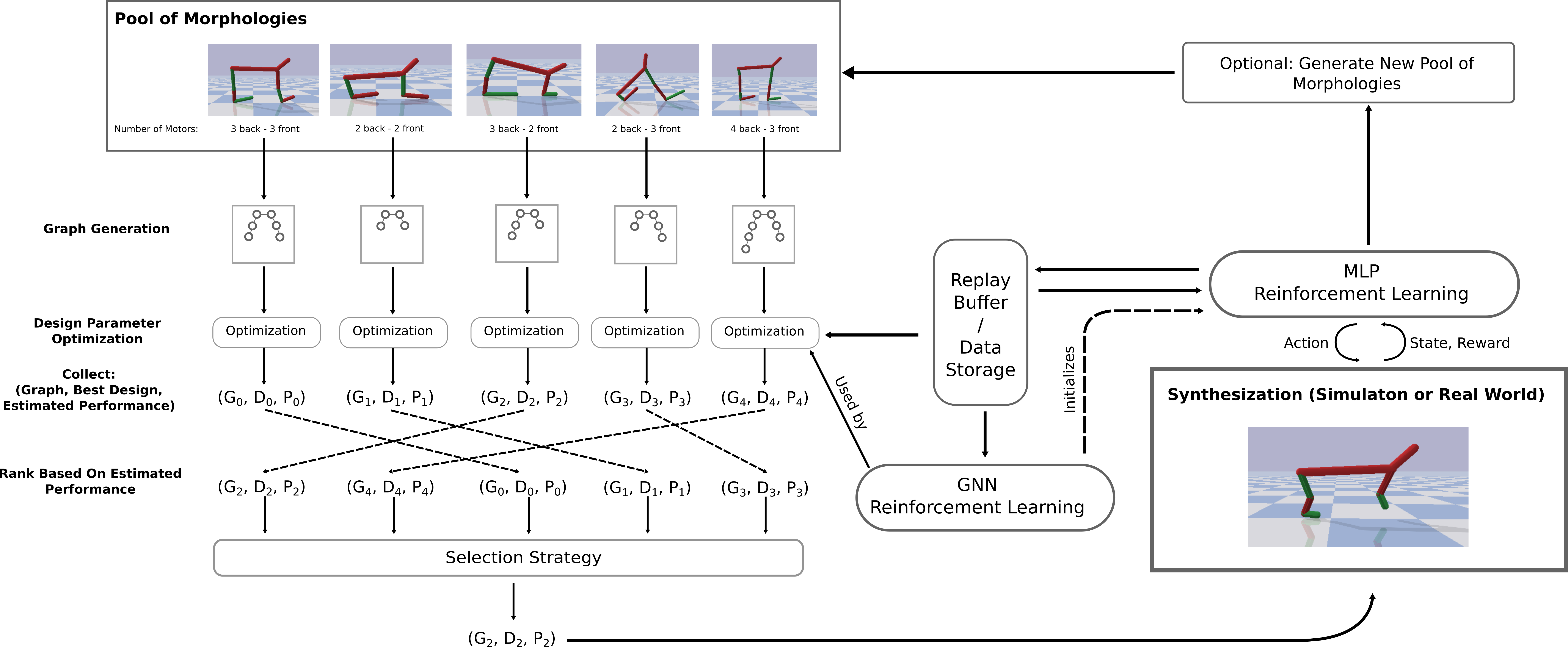}
    \caption{Visualization of the SG-MOrph method described in Algorithm 1.}
    \label{Appendix::Fig::Overview}
\end{figure*}

\paragraph{Overview: }
Figure \ref{Appendix::Fig::Overview} shows a more detailed overview of the proposed SG-MOrph method as a graphical representation of Algorithm 1. 
From our current pool of morphologies, shown at the top left of Figure \ref{Appendix::Fig::Overview}, we first extract the graph representation of each robot morphology considered. 
Practically, these graphs are generated using the Deep Graph Library \cite{wang2019deep}. 
After this, we identify the optimal design parameters for each graph structure using continuous optimization algorithms and the introduced objective function from Eq.\ \ref{Eq::mopt}. More details for this optimization step are provided in the next paragraph. 
Each optimization returns us, for a graph structure $\klgraph$, the optimal design variables $\kldesign$ and its \textit{estimated performance} from the objective function. 
These triples $(\klgraph_i, \kldesign_i, P_i)$ are now ranked based on their performance $P_i$. 
Using a selection strategy, we select a single morphology-design combination from this ranked set. 
This selection strategy can either be greedy, by selecting always the best performing triple, or by assigning higher selection-probabilities to better performing combinations to allow for some exploration. 
In our implementation we use a CMA-ES inspired selection strategy where the probability of selecting a triple is proportional to $\log(k + 2) - \log(r)$, with $r$ being the rank of the triple and $k$ the total number of graph structures. 

Using the selected morphology-design combination $(\klgraph_i, \kldesign_i)$ we can now synthesize our agent by creating its simulation or producing a prototype in the real world using the required manufacturing methods. 
The optimal behavioural policies for the selected agent are trained using MLP networks with fast inference times and using the Soft Actor Critic (SAC) algorithm. 
To avoid having to start the behavioural learning process from scratch every time a new agent is synthesized, we pre-train the actor and critic MLP networks. 
For this we use a transfer-learning approach for which we query the GNN actor and critic networks for its predicted Q-values and actions given this new, previously unseen, morphology. 
Of course, at the time of the pre-training procedure no data has yet been collected on the new agent morphology. Therefore, we create a dataset of $200,000$ \textit{imagined states}. These states could be in principle generated from a distribution $p(\mathbf{s}\vert \klgraph, \kldesign)$. 
However, in practice we can leverage the data collected on previous agents which are stored in the format $(\mathbf{s}_{i}, \mathbf{a}_i, \mathbf{r}_i, \mathbf{s}_{i+1}, \klgraph_j, \kldesign_j)$ in the global replay buffer for agent morphologies $(\klgraph_j, \kldesign_i)$. 
As long as the graph structure, \ie the number of degrees-of-freedom, are similar to the current morphology, we can simply replace the graph and design parameters and use this data to query the GNN for values in respect to the current agent. 
If the a graph structure has never been encountered before, one can resort to create synthetic data from $p(\mathbf{s}\vert \klgraph, \kldesign)$. 
In our experiments, we will only consider a fixed number of graph structures with different degrees-of-freedom/number of nodes, thus we can rely on data available in the data storage. 
We deem this to be of higher interest for us, since the actual number of motor placements on a real legged robot would be limited as well, limited to the choices of adding or removing a limb-segment on a leg. 
This choice, however, does not limit the number of design choices, such that the number of possible morphology-design combinations is still infinite. 

After the pre-training of the MLP actor and critic is concluded, we continue with the normal reinforcement leaning procedure on the synthesized agent. 
During this phase, one has to train only the MLP networks with SAC. However, in our setup in simulation we train GNN and MLP networks in general in parallel at the end of each episode for a fixed number of iterations. 
The advantage of this approach is in principle that MLP networks have a suitably fast inference time which makes them applicable to locomotion tasks requiring high-frequency control. 
Because our evaluation here was conducted in simulation, we are also able to test the performance of the GNN actors, ignoring their considerably higher inference times stemming from the Message-Passing routine. However, only experience generated by the MLP networks was forwarded and stored in the global replay buffer. 

After training has been concluded on the current prototype, one can optionally update the pool of candidate-graphs using the computed ranking. 
For this, any evolutionary or other discrete mutation-optimization algorithm can be used. 
Finally, we proceed with another round of SG-MOrph. 

\paragraph{Continuous Design Optimization:}
\begin{wrapfigure}{r}{0.5\textwidth}
    \centering
    \includegraphics[width=0.5\textwidth]{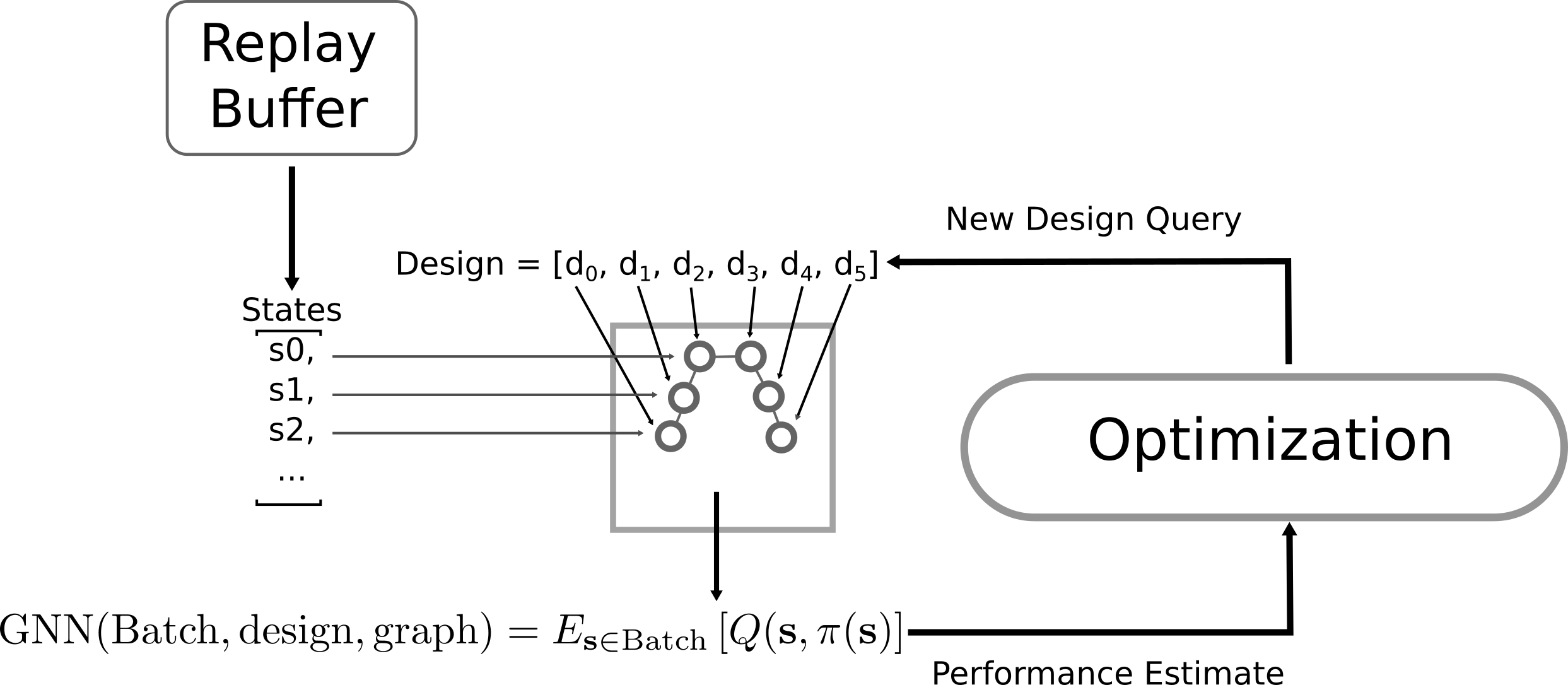}
    \caption{A visualization of the design optimization process for continuous design variables $\kldesign$ given a discrete graph structure $\klgraph$.}
    \label{Appendix::Fig::Optim}
\end{wrapfigure}
In this section we shed further light on the process of optimizing the continuous design parameters given a discrete graph structure, or morphology, $\klgraph$. 
Given a graph $\klgraph$ we find the optimal set of continuous design parameters by optimizing the objective function given in Eq.\ \ref{Eq::mopt}. 
When the continuous optimization algorithm queries the objective function for a design $\kldesign$, we will first set the design parameters at the corresponding nodes for each limb/actuator in the graph $\klgraph$. 
Then, we sample a batch of start states $\mathbf{s}_0$ either from a distribution $p(\mathbf{s}_0 \vert \klgraph, \kldesign)$ or from the replay buffers containing start states from the same graph structures $\klgraph$ (but different design $\kldesign$). 
In our experiments we chose the latter strategy. 
Using the GNN critic and actor we can now query the Q-Value function for the expected performance of a new design $\kldesign$. 
This process is visualized in Figure \ref{Appendix::Fig::Optim}. 
It is worth to note that this objective function is able to provide gradients and is differentiable with respect to the design variables $\kldesign$. 
In practice, we use Particle Swarm Optimization (PSO), exploiting its ability to find global optima, for the optimization of this function. 
This is made possible because the evaluation of this objective is several magnitudes faster than simulation and especially practical in real world scenarios. 
However, if it is possible to reduce computational effort by using gradient-descent based optimization algorithms like stochastic gradient descent, although this will come with a reduced ability to find global optima. 
We execute the PSO algorithm for 50 iterations and 80 particles. 

\paragraph{Graph Structure:} The structure of graphs is largely based on prior work such as \cite{wang2018nervenet, huang2020one}. Here, we have a one-to-one mapping between graph nodes and actuators/motors with an attached limb. 
This means, the number of nodes is equal to the degrees-of-freedom of a robot. 
Each node $n_i$ has three fields: The continuous design parameters $\kldesign^i$, the global state $\mathbf{s}^g$ and the actuator/limb state $\mathbf{s}^i$. 
The design parameters $\kldesign^i \in \mathbb{R}^d$ of a node describe the motor or limb properties such as force, length, material mixing-ratio or the orientation of the limb. 
The global state $\mathbf{s}^g$ contains all state information which is not specific to any node, such as the velocity, orientation or position information of the center of gravity of a robot. 
While prior work experimented with graphs where such information would be assigned to a body special node, we found that assigning this information to all nodes lead to a more reliable and better performance. 
Furthermore, it enables us to use a single set of shared networks between all nodes. Otherwise, using a special node processing only global state information would require an extra set of processing networks due to the different state sizes. 
Finally, the node state $\mathbf{s}^i$ contains only state information of the corresponding motor or limb. 
In our experiments, this state usually contains state information such as the joint position or velocity.

\subsection{Hyperparameters}
The hyperparameters used in experiments can be found in Table \ref{Appendix::Table::Hyper}. The selection of the hyper-parameters for MLP and Graph Neural Networks is based on parameters used in prior work \cite{luck2020data, wang2018nervenet, wang2018neural, huang2020one} and were determined in preliminary experiments.

\begin{table}[h!]
\caption{Hyperparameters of the used Specialist and Generalist networks in SG-MOrph. Table (a) shows the hyperparameters used in standard SAC for $N$ action dimensions. For Graph Neural Networks or Graph-SAC the parameters in Table (b) were used. Finally, Table (c) lists parameters used during the transfer learning process from GNN to MLP networks in SG-MOrph.}
\label{Appendix::Table::Hyper}
\begin{subtable}{.5\linewidth}
    \centering
    \caption{(MLP-) SAC}
\begin{tabular}{lc}
Discount               & $0.975$           \\ \rowcolor[gray]{0.925}
Soft Target Tau        & $0.01$            \\ 
Actor LR               & $0.001$           \\ \rowcolor[gray]{0.925}
Critic LR              & $0.001$           \\
$\alpha$               & $\frac{0.01}{N}$  \\ \rowcolor[gray]{0.925}
Batch Size             & $256$             \\
Hidden Layers          & $(200, 200, 200)$ \\ \rowcolor[gray]{0.925}
Activations            & ReLU              \\
Optimizer              & Adam              \\ \rowcolor[gray]{0.925}
Iterations per Episode & $1000$    \\       
\end{tabular}
    \vspace{12ex}
\end{subtable}%
\begin{subtable}{.5\linewidth}
    \centering
    \caption{Graph-SAC}
\begin{tabular}{lc}
Discount                    & $0.975$              \\ \rowcolor[gray]{0.925}
Soft Target Tau             & $0.01$               \\
Actor LR                    & $0.0005$             \\ \rowcolor[gray]{0.925}
Critic LR                   & $0.0005$             \\
$\alpha$                    & $0.01$               \\ \rowcolor[gray]{0.925}
Batch Size                  & $128 \times 5 = 640$ \\
Batch: Nmbr of Graphs       & 5                    \\ \rowcolor[gray]{0.925}
Hidden Layers               & $(256,256)$          \\
Message Size                & 64                   \\ \rowcolor[gray]{0.925}
Hidden State Size           & 64                   \\
Activation                  & ReLU                 \\ \rowcolor[gray]{0.925}
Message/Hidden State Activ. & Tanh                 \\
Optimizer                   & Adam                 \\ \rowcolor[gray]{0.925}
Iterations per Episode      & 100                  \\
Message Passing Rounds      & 4                   
\end{tabular}
\end{subtable}
\vspace{3ex}
\begin{subtable}{\linewidth}
    \centering 
    \caption{Transfer Learning}
\begin{tabular}{lc}
Iterations             & $20,000$  \\ \rowcolor[gray]{0.925}
Actor LR               & $0.0005$  \\
Critic LR              & $0.0005$  \\ \rowcolor[gray]{0.925}
Batch Size             & 256       \\
Generated Dataset Size & $200,000$
\end{tabular}
\end{subtable}
\end{table}

\subsection{Environment 1: HalfCheetah}
The goal for the HalfCheetah task was to provide an environment with a variety of agents with different degrees-of-freedom resembling the original HalfCheetah task provided in MuJoCo and PyBullet environments as closely as possible. 
While the original HalfCheetah task in PyBullet has a delicately configured model with individual actuator properties for each joint, we will simplify this and use the same type of actuator for each joint. 
The motors in our HalfCheetah model will have $65Nm$, a movement range of $0.6$ radians, a damping factor of $4.5$ and a stiffness of $180$. 
Actions, \ie motor commands, are continuous values in the range of $[-1,1]$ and episodes are $600$ time steps long. 
In this task we will consider up to six different graph structures, with design parameters representing the length of actuated leg limbs in the range of $[0.2,0.6]$. 
The state space has five global state variables $\mathbf{s}^g$, which represent the normalized height, x-velocity, pitch, pitch velocity and z-velocity of the main torso. 
Individual node states $\mathbf{s}^i$ are two-dimensional and contain joint position and joint velocity information, but no further positional or velocity information in the world coordinate frame. 
The reward function is based on the x-velocity of the torso and is given by $r(\mathbf{s}) = \max(\frac{\Delta x}{10},0)$.

\begin{figure*}[h!]
    \centering
    \begin{subfigure}[t]{0.25\textwidth}
        \centering
        \includegraphics[width=\textwidth]{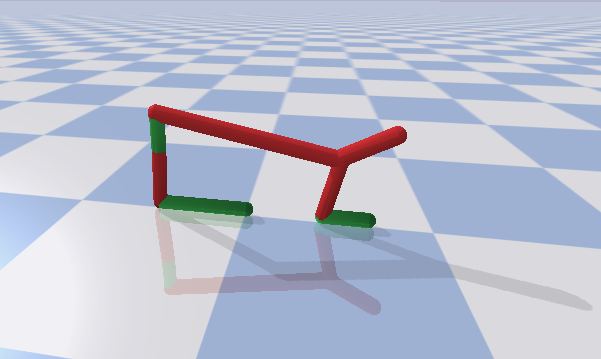}
        \caption{5 DoF}
    \end{subfigure}%
    ~ 
    \begin{subfigure}[t]{0.25\textwidth}
        \centering
        \includegraphics[width=\textwidth]{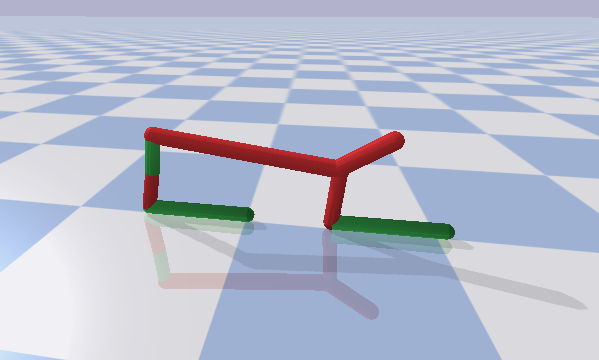}
        \caption{5 DoF}
    \end{subfigure}%
    ~
    \begin{subfigure}[t]{0.25\textwidth}
        \centering
        \includegraphics[width=\textwidth]{fig2/designs/best/hc/hc_2_cut.png}
        \caption{5 DoF}
    \end{subfigure}%
    \caption{Selection of optimized morphology-design-combinations found for HalfCheetah.}
    \label{Appendix::Fig::HC::optimal}
\end{figure*}

\begin{figure*}[h!]
    \centering
    \begin{subfigure}[t]{0.25\textwidth}
        \centering
        \includegraphics[width=\textwidth]{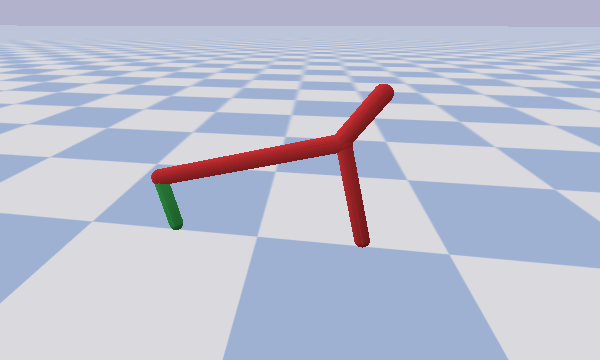}
        \caption{Morph. 1, Design 1}
    \end{subfigure}%
    ~ 
    \begin{subfigure}[t]{0.25\textwidth}
        \centering
        \includegraphics[width=\textwidth]{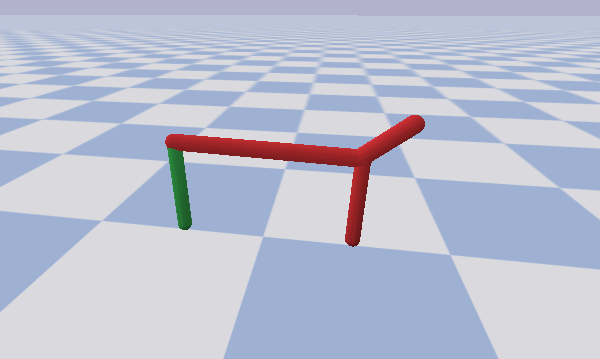}
        \caption{Morph. 1, Design 2}
    \end{subfigure}%
    ~
    \begin{subfigure}[t]{0.25\textwidth}
        \centering
        \includegraphics[width=\textwidth]{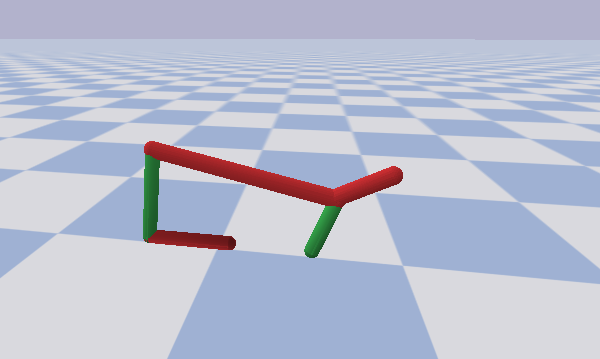}
        \caption{Morph. 2, Design 1}
    \end{subfigure}%
    ~ 
    \begin{subfigure}[t]{0.25\textwidth}
        \centering
        \includegraphics[width=\textwidth]{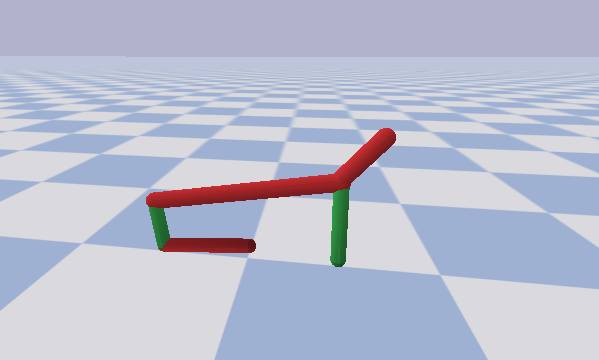}
        \caption{Morph. 2, Design 2}
    \end{subfigure}
    
    \begin{subfigure}[t]{0.25\textwidth}
        \centering
        \includegraphics[width=\textwidth]{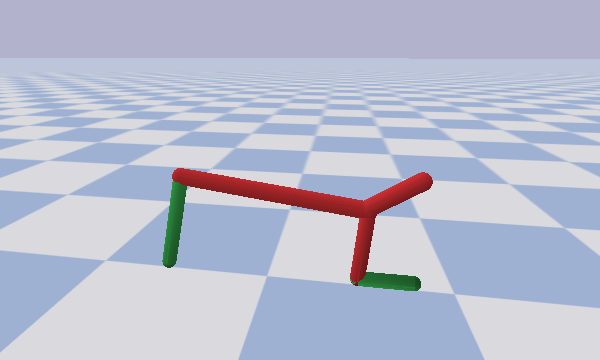}
        \caption{Morph. 3, Design 1}
    \end{subfigure}%
    ~ 
    \begin{subfigure}[t]{0.25\textwidth}
        \centering
        \includegraphics[width=\textwidth]{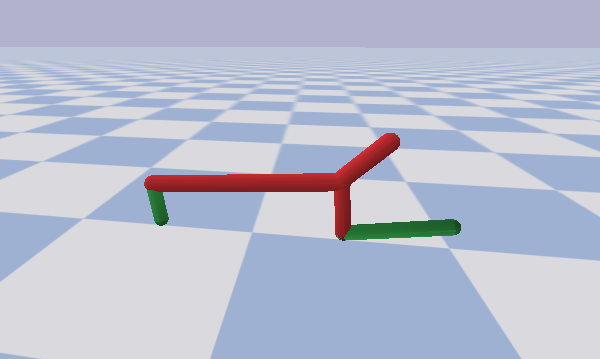}
        \caption{Morph. 3, Design 2}
    \end{subfigure}%
    ~
    \begin{subfigure}[t]{0.25\textwidth}
        \centering
        \includegraphics[width=\textwidth]{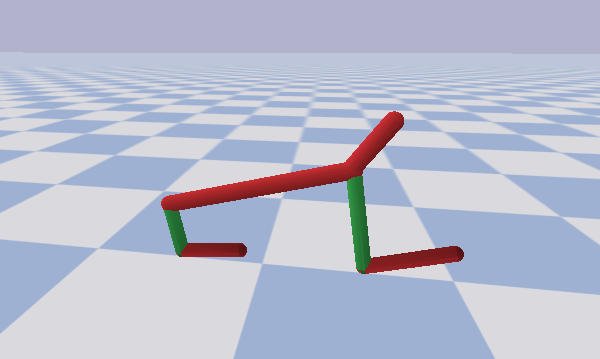}
        \caption{Morph. 4, Design 1}
    \end{subfigure}%
    ~ 
    \begin{subfigure}[t]{0.25\textwidth}
        \centering
        \includegraphics[width=\textwidth]{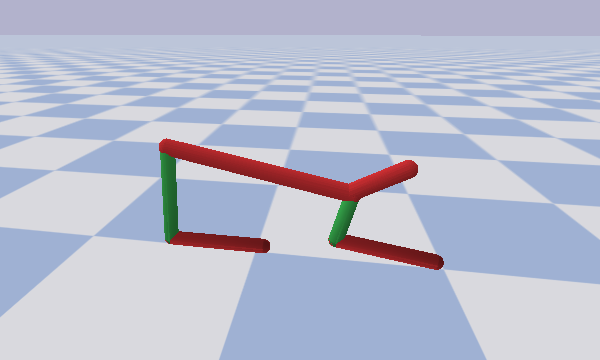}
        \caption{Morph. 4, Design 2}
    \end{subfigure}
    
    \begin{subfigure}[t]{0.25\textwidth}
        \centering
        \includegraphics[width=\textwidth]{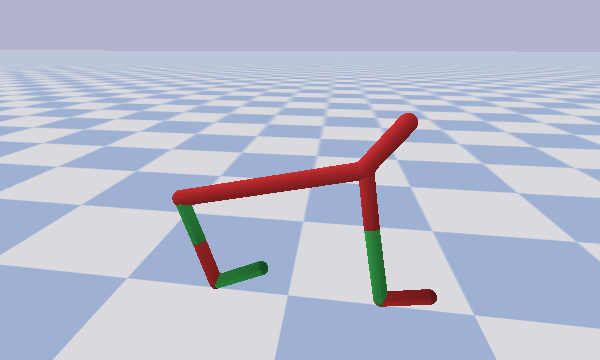}
        \caption{Morph. 5, Design 1}
    \end{subfigure}%
    ~ 
    \begin{subfigure}[t]{0.25\textwidth}
        \centering
        \includegraphics[width=\textwidth]{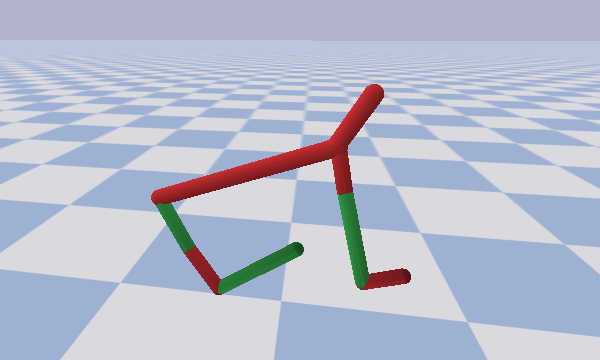}
        \caption{Morph. 5, Design 2}
    \end{subfigure}%
    ~
    \begin{subfigure}[t]{0.25\textwidth}
        \centering
        \includegraphics[width=\textwidth]{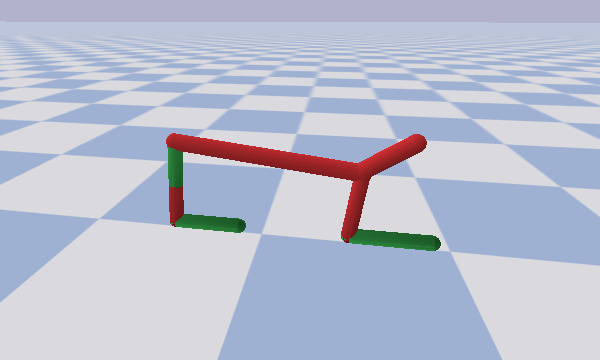}
        \caption{Morph. 6, Design 1}
    \end{subfigure}%
    ~ 
    \begin{subfigure}[t]{0.25\textwidth}
        \centering
        \includegraphics[width=\textwidth]{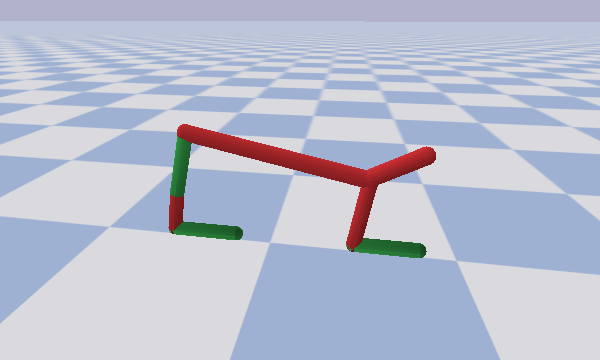}
        \caption{Morph. 6, Design 2}
    \end{subfigure}
    
    \begin{subfigure}[t]{0.25\textwidth}
        \centering
        \includegraphics[width=\textwidth]{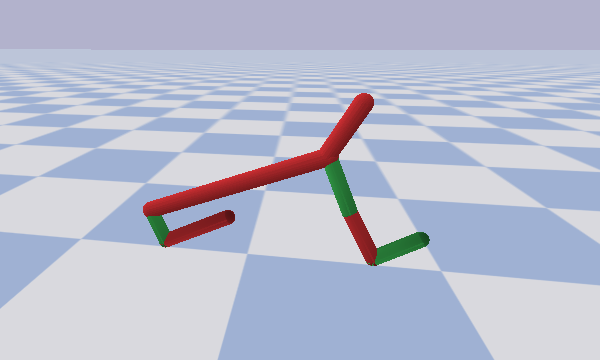}
        \caption{Morph. 7, Design 1}
    \end{subfigure}%
    ~ 
    \begin{subfigure}[t]{0.25\textwidth}
        \centering
        \includegraphics[width=\textwidth]{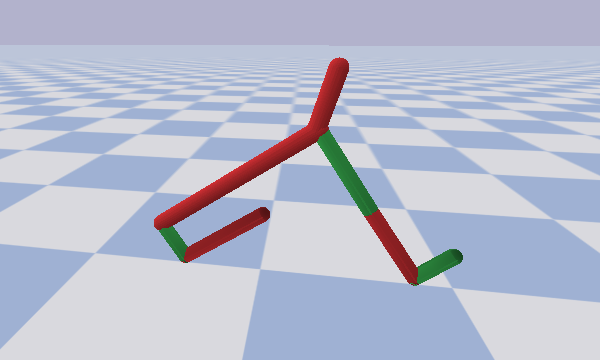}
        \caption{Morph. 7, Design 2}
    \end{subfigure}
    \caption{Initial designs pool for HalfCheetah consisting of seven different morphologies, \ie graph structures, ranging from two to six degrees-of-freedom.
    Design variables for these initial designs were selected from a uniform probability distribution. These morphology-design combinations are always trained on at the beginning of the design optimization process. 
    }
    \label{Appendix::Fig::HC::Initial}
\end{figure*}

\begin{figure}[h!]
    \centering
    \includegraphics[width=0.8\textwidth]{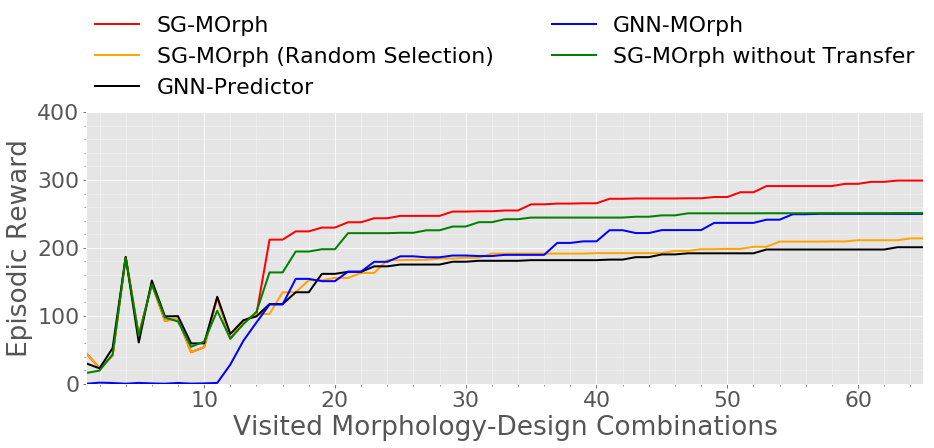}
    \caption{
    This figure shows the performance of the current best design uncovered after $x$ design optimization rounds, averaged over ten experiments.
    In each experiment, we perform 25 morphology optimizations and 25 selections of random morphology-design combinations, after training on the initial design pool. 
    GNN policies are trained once the replay buffer contains the majority of initial designs, to prevent early convergence on one graph structure. For GNN-MOrph, which only executed GNN policies on agents, the initial designs were trained with MLP networks to allow for a better and fairer comparison to other baselines, hence GNN policies showing an episodic reward of 0 for the first few morphologies. }
    \label{Appendix::Fig::HC::compbad}
\end{figure}

The fixed set of randomly selected initial designs used in experiments can be found in Figure \ref{Appendix::Fig::HC::Initial}. Figure \ref{Appendix::Fig::HC::optimal} shows a selection of three optimal designs found for the HalfCheetah environment. 
In addition to the results presented in the main paper we show here a further analysis of the performance of SG-MOrph versus baselines. 
Figure \ref{Appendix::Fig::HC::compbad} shows the best performance of created morphologies for each algorithm, averaged over ten experiments. 
The performances shown for the initial 14 morphologies shown are the same for all algorithms because they belong to the initial set of morphology-design combinations shown in Fig.\ \ref{Appendix::Fig::HC::Initial}. 
Thereafter, all algorithms perform 25 design optimization and 25 random design selections resulting in a total of 64 simulated agents.
SG-MOrph is able to find early on good performing morphologies and outperforms baseline approaches. 
Using only GNNs as well as using SG-MOrph without transfer learning converges to the same level of performance after roughly 54 visited morphology-design combinations. 
Using Graph Neural Networks to predict performances solely on graph structure and design parameters, as proposed in \cite{wang2018neural}, does not outperform the random baseline. 
As discussed in the main paper, the GNN-prediction network can only leverage one datum per morphology-design combination, the best performance, while the GNN-critic is able to leverage $65,000$ samples. 
We see this as the main reason for its bad performance in this training regime. Previous approaches trained such prediction networks on larger datasets due to the ability to simulate large populations in parallel.

\begin{figure*}[h!]
    \centering
    \begin{subfigure}[t]{0.5\textwidth}
        \centering
        \includegraphics[width=\textwidth]{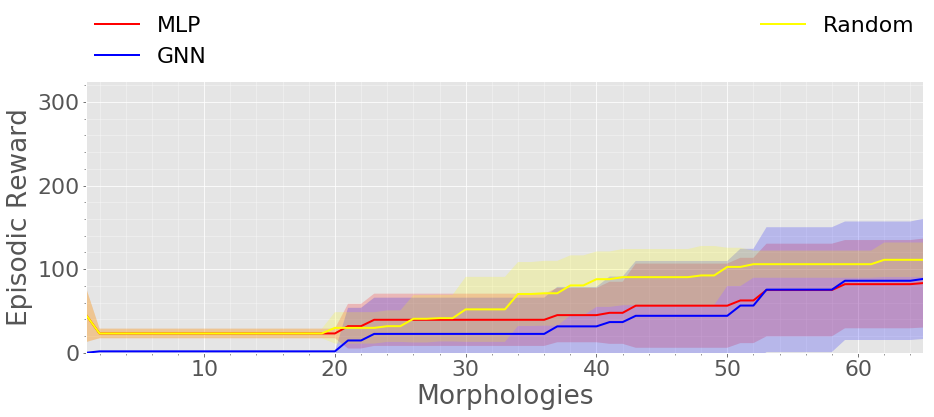}
        \caption{Morphology 1}
    \end{subfigure}%
    ~ 
    \begin{subfigure}[t]{0.5\textwidth}
        \centering
        \includegraphics[width=\textwidth]{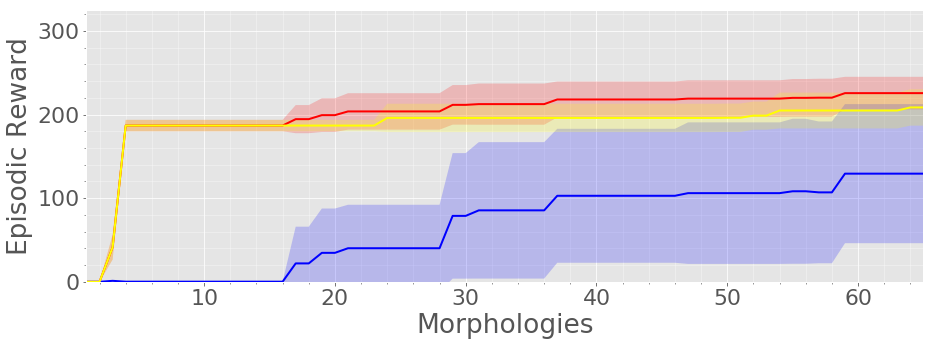}
        \caption{Morphology 2}
    \end{subfigure}
    
    \begin{subfigure}[t]{0.5\textwidth}
        \centering
        \includegraphics[width=\textwidth]{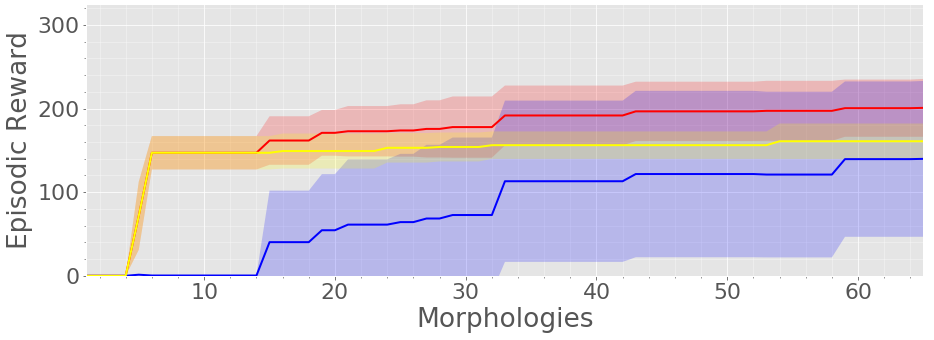}
        \caption{Morphology 3}
    \end{subfigure}%
    ~ 
    \begin{subfigure}[t]{0.5\textwidth}
        \centering
        \includegraphics[width=\textwidth]{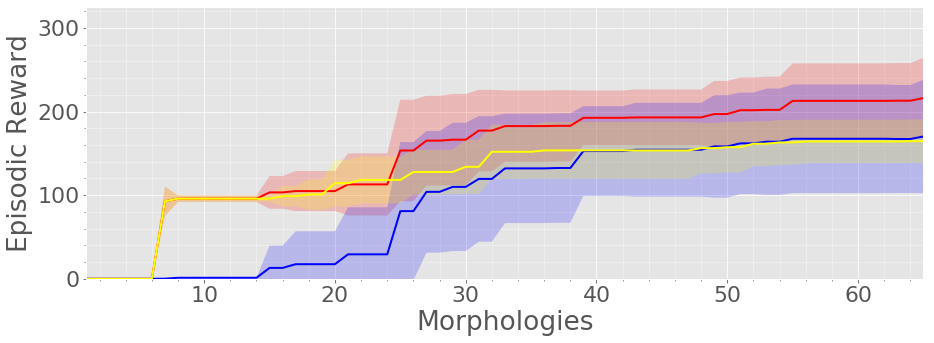}
        \caption{Morphology 4}
    \end{subfigure}
    
    \begin{subfigure}[t]{0.5\textwidth}
        \centering
        \includegraphics[width=\textwidth]{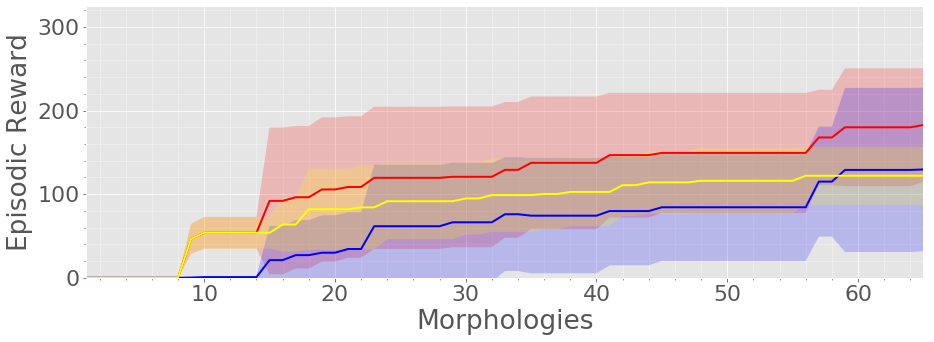}
        \caption{Morphology 5}
    \end{subfigure}%
    ~ 
    \begin{subfigure}[t]{0.5\textwidth}
        \centering
        \includegraphics[width=\textwidth]{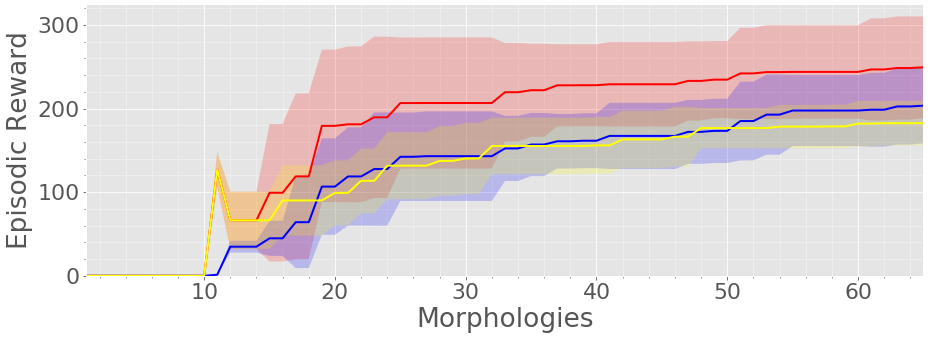}
        \caption{Morphology 6}
    \end{subfigure}
    
    \begin{subfigure}[t]{0.5\textwidth}
        \centering
        \includegraphics[width=\textwidth]{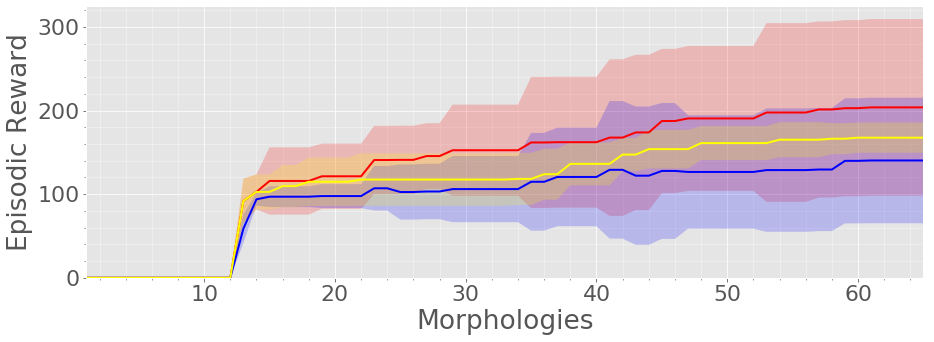}
        \caption{Morphology 7}
    \end{subfigure}
    \caption{Performance of SG-MOrph on the HalfCheetah environment. Graphs show the average and standard deviation of the best performance seen in 100 episodes for each morphology-design combination, computed from ten experiments. The first 14 morphologies are part of the fixed initial design pool (Fig. \ref{Appendix::Fig::HC::Initial}) and trained on sequentially, thereafter 25 design optimization and 25 random design selections are performed. }
    \label{Appendix::Fig::HC::sgmorphdetail}
\end{figure*}

Finally, Figure \ref{Appendix::Fig::HC::sgmorphdetail} shows the performance of optimized designs vs.\ randomly selected designs. Furthermore, for evaluation purposes we also compared the performance of the MLP policies vs.\ the GNN policies. 
We found that in this task SG-MOrph is able to improve upon the initially selected random designs, chosen from a random uniform distribution, and that the high-frequency MLP policy is able to adapt much faster to new morphology-design combinations than the GNN policy. 
Furthermore, only in the morphology with the smallest number of degrees-of-freedom, morphology 1 with two degrees-of-freedom, randomly selecting designs performs better than optimizaton. 
We find that there are two reasons for this: First, due to the small number of degrees-of-freedom the number of design parameters is also small. Since we optimize only the length of limbs, HalfCheetah has in this case only two design dimensions. 
For such a small space randomly selecting designs is a reasonable strategy. However, once we start increasing the number of DoFs and thus increase the number of dimensions of the design space, the curse of dimensionality starts to take effect making it much harder to cover the volume of the space via random sampling. 
Secondly, this morphology shows also the worst performance. Hence, due to the probabilistic selection strategy applied in SG-MOrph, the algorithm will select this morphology for optimization only with a very small probability.

\begin{figure}
    \centering
    \begin{subfigure}[t]{\textwidth}
        \centering
        \includegraphics[width=0.8\textwidth]{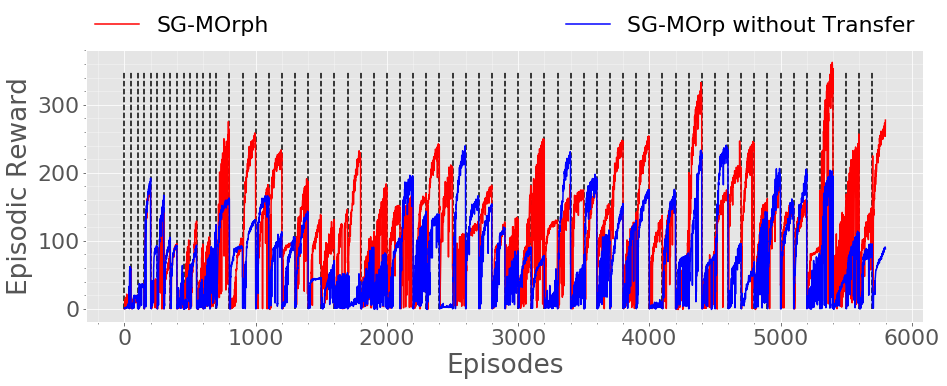}
    \caption{Raw episodic reward data from two runs with SG-MOrph and SG-MOrph not using the proposed transfer learning approach. Dotted vertical lines indicate the switch between two morphology-design-combinations.}
    \end{subfigure}
    
    \begin{subfigure}[t]{\textwidth}
        \centering
        \includegraphics[width=0.8\textwidth]{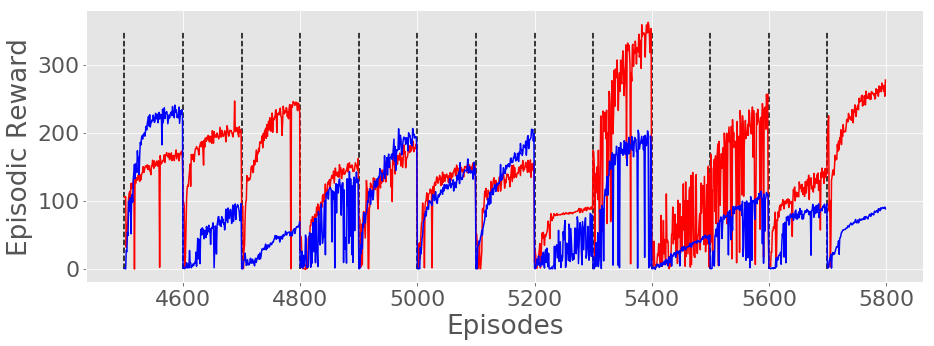}
        \caption{Detail view of Fig.\ (a), showing the last 13 morphologies trained on.}
    \end{subfigure}
    \caption{This figure presents a detailed view of the episodic rewards collected during a typical run of SG-MOrph. Dashed vertical lines indicate the switch between two morphology-design combinations. This plot includes both optimized as well as randomly selected designs. As a reminder, after the initial design were visited, SG-MOrph executes 25 design optimizations and 25 random design selections in alternating fashion. These graphs show the impact of the used transfer learning approach, which allows SG-MOrph to quickly reach high performances in the first few episodes spent on a new agent. If no transfer learning is used, SG-MOrph is more likely to require more episodes to converge to higher episodic rewards. }
    \label{Appendix::Fig::HC::raw}
\end{figure}

Figure \ref{Appendix::Fig::HC::raw} shows exemplary two full executions of SG-MOrph with and without the proposed transfer learning approach. 
It can be seen that using the pre-training mechanism allows the MLP networks in SG-MOrph more often to find good initial behaviours during the first few episodes. 

\clearpage
\subsection{Environment 2: Crawler}
\begin{figure*}[h!]
    \centering
    \begin{subfigure}[t]{0.25\textwidth}
        \centering
        \includegraphics[width=\textwidth]{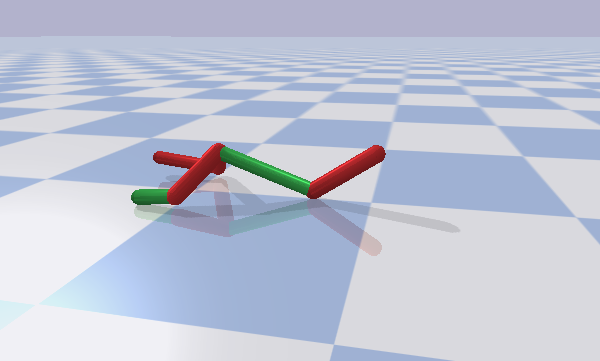}
        \caption{4 DoF}
    \end{subfigure}%
    ~ 
    \begin{subfigure}[t]{0.25\textwidth}
        \centering
        \includegraphics[width=\textwidth]{fig2/designs/best/walker/walker_1_cut.png}
        \caption{6 DoF}
    \end{subfigure}%
    ~
    \begin{subfigure}[t]{0.25\textwidth}
        \centering
        \includegraphics[width=\textwidth]{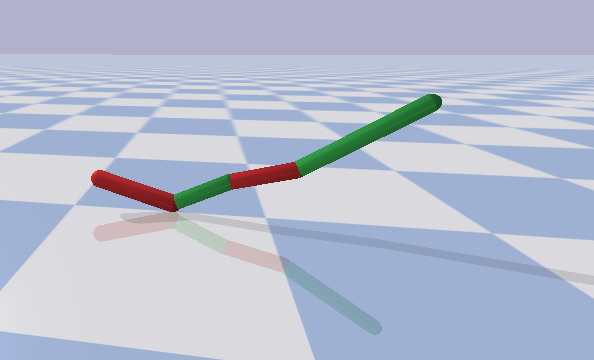}
        \caption{3 DoF}
    \end{subfigure}%
    \caption{Selection of optimized  morphologies found for Crawler.}
    \label{Appendix::Fig::Crawler::Optimal}
\end{figure*}

\begin{figure*}[h!]
    \centering
    \begin{subfigure}[t]{0.25\textwidth}
        \centering
        \includegraphics[width=\textwidth]{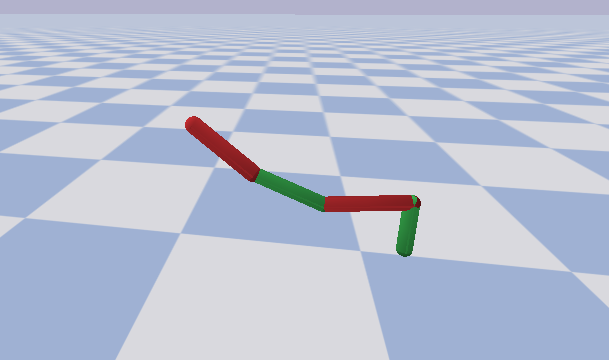}
        \caption{Morph. 1, Design 1}
    \end{subfigure}%
    ~ 
    \begin{subfigure}[t]{0.25\textwidth}
        \centering
        \includegraphics[width=\textwidth]{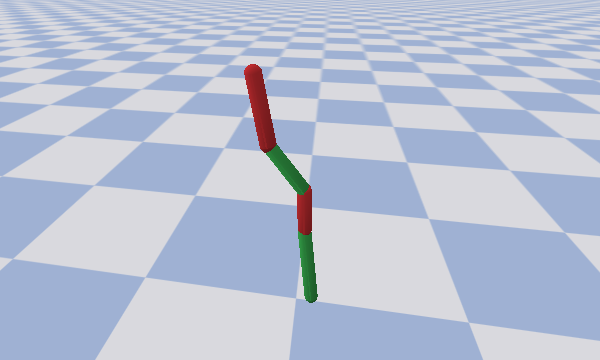}
        \caption{Morph. 1, Design 2}
    \end{subfigure}%
    ~
    \begin{subfigure}[t]{0.25\textwidth}
        \centering
        \includegraphics[width=\textwidth]{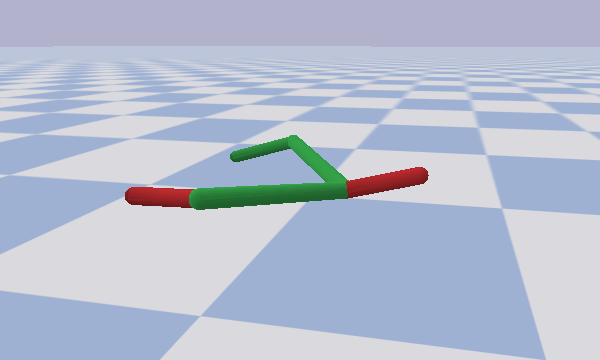}
        \caption{Morph. 2, Design 1}
    \end{subfigure}%
    ~ 
    \begin{subfigure}[t]{0.25\textwidth}
        \centering
        \includegraphics[width=\textwidth]{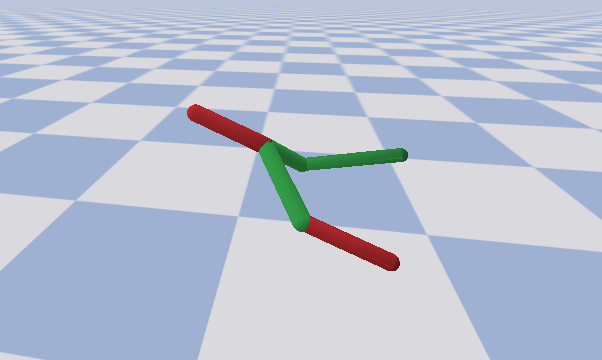}
        \caption{Morph. 2, Design 2}
    \end{subfigure}
    
    \begin{subfigure}[t]{0.25\textwidth}
        \centering
        \includegraphics[width=\textwidth]{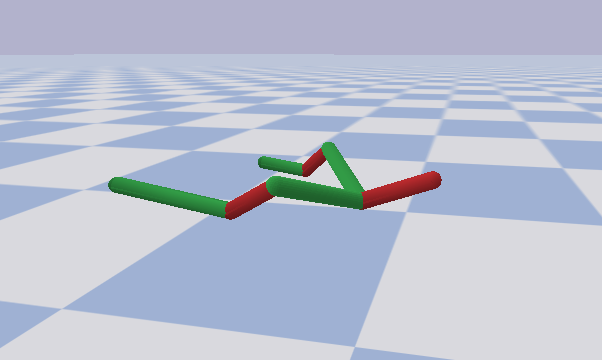}
        \caption{Morph. 3, Design 1}
    \end{subfigure}%
    ~ 
    \begin{subfigure}[t]{0.25\textwidth}
        \centering
        \includegraphics[width=\textwidth]{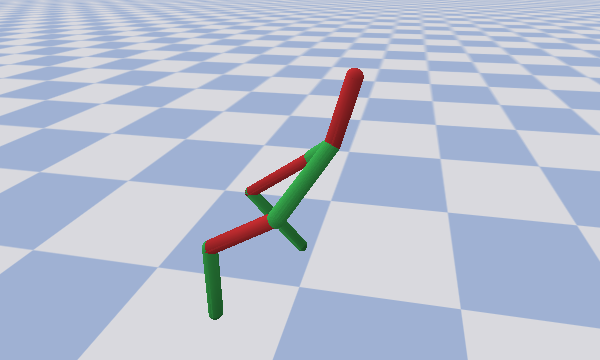}
        \caption{Morph. 3, Design 2}
    \end{subfigure}%
    ~
    \begin{subfigure}[t]{0.25\textwidth}
        \centering
        \includegraphics[width=\textwidth]{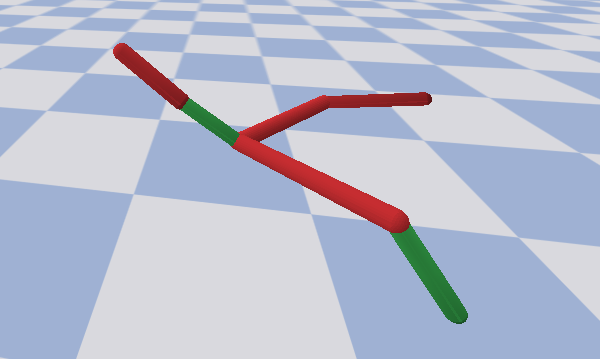}
        \caption{Morph. 4, Design 1}
    \end{subfigure}%
    ~ 
    \begin{subfigure}[t]{0.25\textwidth}
        \centering
        \includegraphics[width=\textwidth]{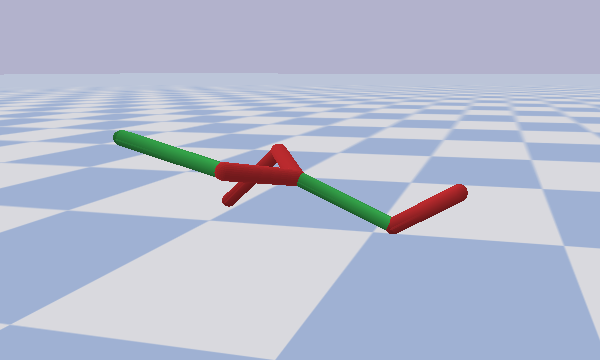}
        \caption{Morph. 4, Design 2}
    \end{subfigure}
    
    \begin{subfigure}[t]{0.25\textwidth}
        \centering
        \includegraphics[width=\textwidth]{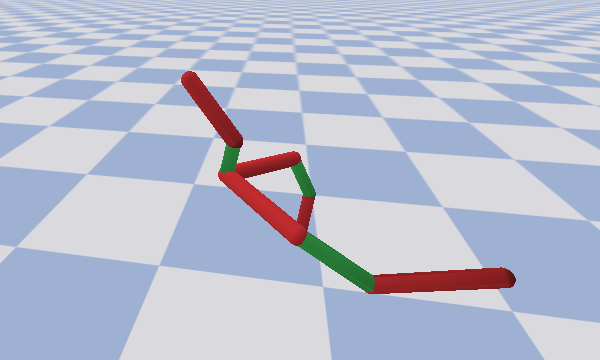}
        \caption{Morph. 5, Design 1}
    \end{subfigure}%
    ~ 
    \begin{subfigure}[t]{0.25\textwidth}
        \centering
        \includegraphics[width=\textwidth]{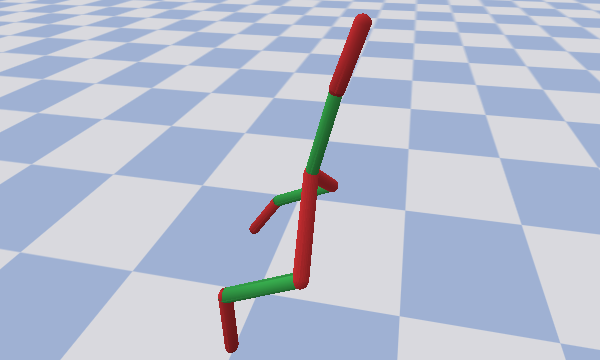}
        \caption{Morph. 5, Design 2}
    \end{subfigure}
    \caption{Initial designs pool for Crawler consisting of five different morphologies, \ie graph structures, ranging from three to six degrees-of-freedom.}
    \label{Appendix::Fig::Crawler::Initial}
\end{figure*}

The crawler task was inspired by the Walker environment in PyBullet. 
However, in practice we found in our experiments that the algorithms preferred a crawling motion over ground due to the added stability. 
If the \textit{head} of the crawler would have a pitch below $1.25$ radians to either side we would stop the episode and assign a negative reward of $-1.0$. 
Otherwise, similarly to HalfCheetah, the agent would receive a reward of $r(\mathbf{s}) = \max(\frac{\Delta x}{10},0)$, with $\Delta x$ being the velocity along the x-axis. 
In this task we extended the number of continuous design parameters to optimize by both optimizing length of limbs as well as their orientation. 
Hence, each limb has two parameters: One length parameter in $[0.2,0.6]$ and an orientation parameter in $[-1.0, 1.0]$ radians. 
Motor properties were similar to HalfCheetah with the exception of the power: We used here $30Nm$ for each actuator, making them weaker. 
This forces the agent to discover more natural movement behaviours and prevents the exploitation of momentum seen in HalfCheetah (see videos). 
We optimized for five different discrete morphological structures, the set of initial morphology-design combinations can be found in Figure \ref{Appendix::Fig::Crawler::Initial}. 
A selection of optimized morphology-design combinations with high episodic rewards is shown in Figure \ref{Appendix::Fig::Crawler::Optimal}. 

\begin{figure*}[h!]
    \centering
    \begin{subfigure}[t]{0.5\textwidth}
        \centering
        \includegraphics[width=\textwidth]{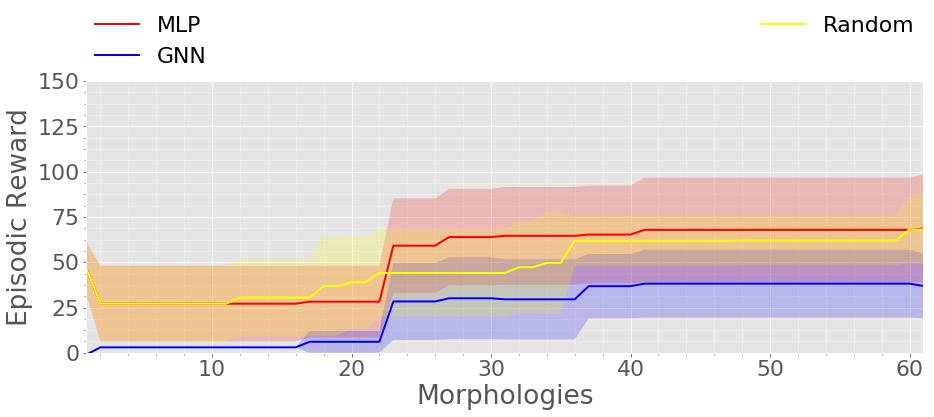}
        \caption{Morphology 1}
    \end{subfigure}%
    ~ 
    \begin{subfigure}[t]{0.5\textwidth}
        \centering
        \includegraphics[width=\textwidth]{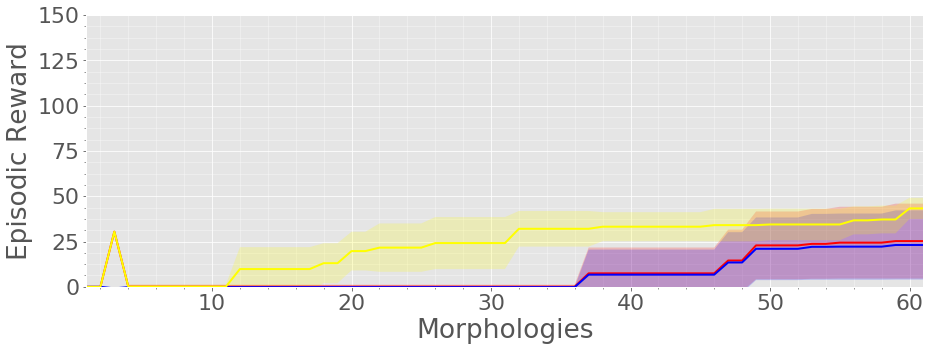}
        \caption{Morphology 2}
    \end{subfigure}
    
    \begin{subfigure}[t]{0.5\textwidth}
        \centering
        \includegraphics[width=\textwidth]{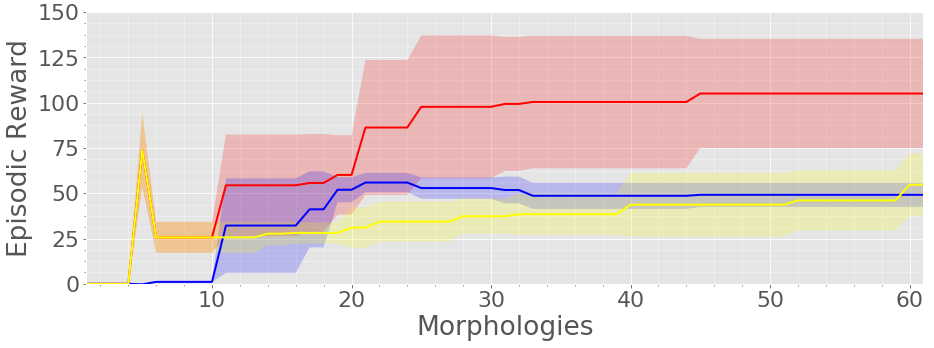}
        \caption{Morphology 3}
    \end{subfigure}%
    ~ 
    \begin{subfigure}[t]{0.5\textwidth}
        \centering
        \includegraphics[width=\textwidth]{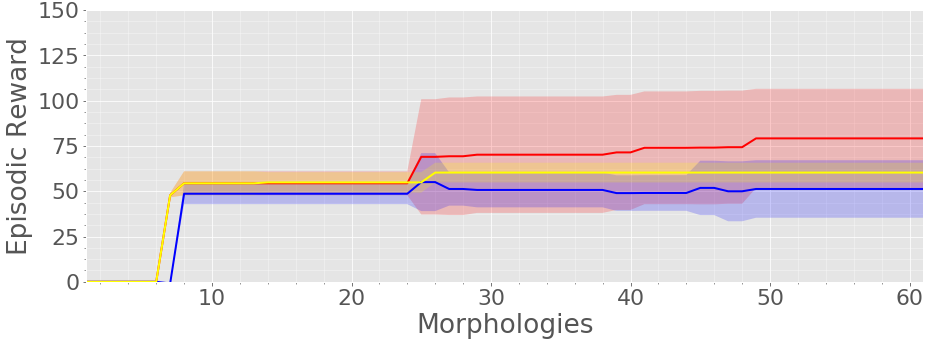}
        \caption{Morphology 4}
    \end{subfigure}
    
    \begin{subfigure}[t]{0.5\textwidth}
        \centering
        \includegraphics[width=\textwidth]{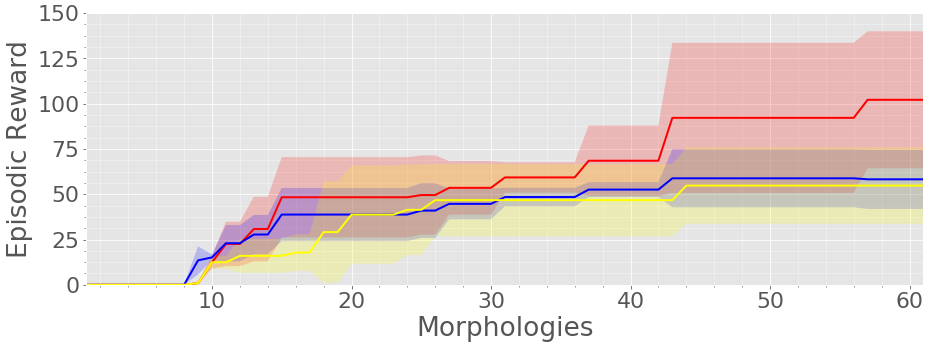}
        \caption{Morphology 5}
    \end{subfigure}%

    \caption{Performance of SG-MOrph on the Crawler environment. Graphs show the average and standard deviation of the best performance seen in 100 episodes for each morphology-design combination, computed from five experiments. The first 10 morphologies/designs belong to the fixed initial design pool (Fig. \ref{Appendix::Fig::Crawler::Initial}), thereafter 25 design optimization and 25 random design selections are performed. Figures a-e show the performance of SG-MOrph split into the five different morphological structures considered.}
    \label{Appendix::Fig::Crawler::morphdata}
\end{figure*}

Figure \ref{Appendix::Fig::Crawler::morphdata} shows the performance of SG-MOrph policies over one run split between the five distinct graph/morphological structures. 
It can be seen that MLP policies are faster at adapting to new designs than their GNN counterparts. 
Similar to HalfCheetah, we see that in the morphology with the worst performance the random baseline, which selects designs and morphologies randomly, outperforms SG-Morph. 
In all other cases, SG-MOrph is able to find both better design parameters and behaviours through the proposed co-adaptation approach. 

\subsection{Environment 3: Multi-Ped}
The Multi-Ped environment contains a range of two-, three- and four-legged robots in different configurations. 
These morphologies, however, do not feature extras such as a head-like fixed appendix at the torso or feet, \ie the last segments being always oriented at 90 degrees. 
Motor parameters are set to the same values as in the HalfCheetah environment, with the exception of the motor force which was set to $30Nm$. 
Reward function and state spaces are the same as in the Crawler environment. 

\begin{figure*}[h!]
    \centering
    \begin{subfigure}[t]{0.25\textwidth}
        \centering
        \includegraphics[width=\textwidth]{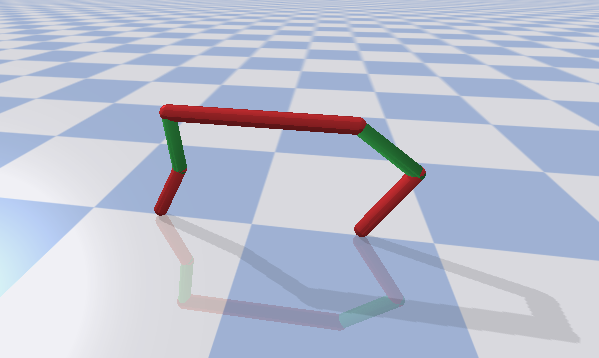}
        \caption{4 DoF}
    \end{subfigure}%
    ~ 
    \begin{subfigure}[t]{0.25\textwidth}
        \centering
        \includegraphics[width=\textwidth]{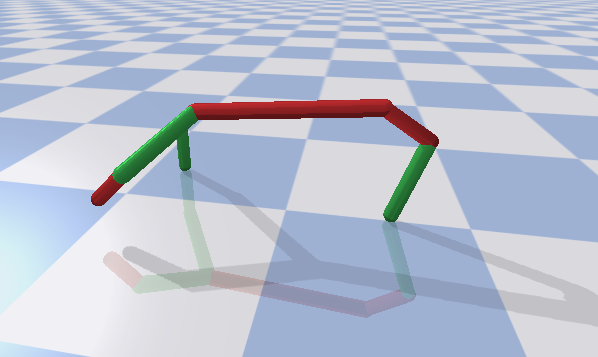}
        \caption{5 DoF}
    \end{subfigure}%
    ~
    \begin{subfigure}[t]{0.25\textwidth}
        \centering
        \includegraphics[width=\textwidth]{fig2/designs/best/legged/legged_2_cut.png}
        \caption{5 DoF}
    \end{subfigure}%
    \caption{Selection of optimized  morphologies found for Multi-Ped.}
    \label{Appendix::Fig::multiped::optimal}
\end{figure*}

\begin{figure*}[h!]
    \centering
    \begin{subfigure}[t]{0.5\textwidth}
        \centering
        \includegraphics[width=\textwidth]{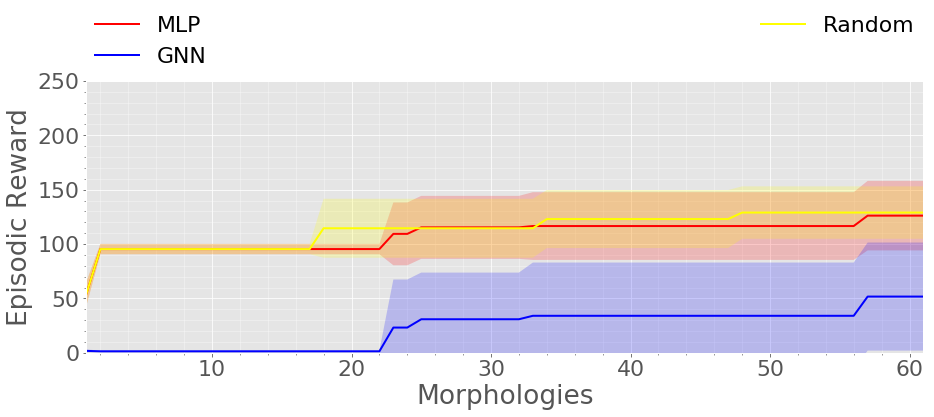}
        \caption{Morphology 1}
    \end{subfigure}%
    ~ 
    \begin{subfigure}[t]{0.5\textwidth}
        \centering
        \includegraphics[width=\textwidth]{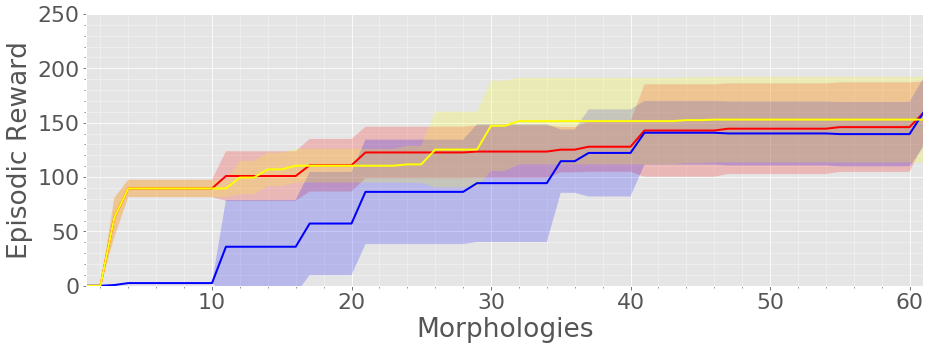}
        \caption{Morphology 2}
    \end{subfigure}
    
    \begin{subfigure}[t]{0.5\textwidth}
        \centering
        \includegraphics[width=\textwidth]{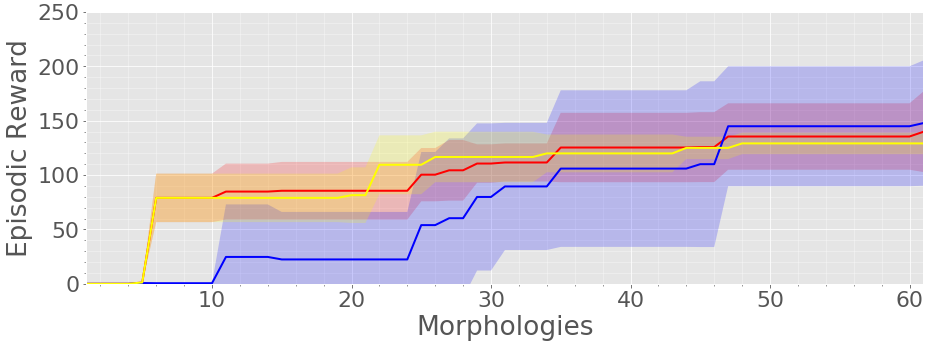}
        \caption{Morphology 3}
    \end{subfigure}%
    ~ 
    \begin{subfigure}[t]{0.5\textwidth}
        \centering
        \includegraphics[width=\textwidth]{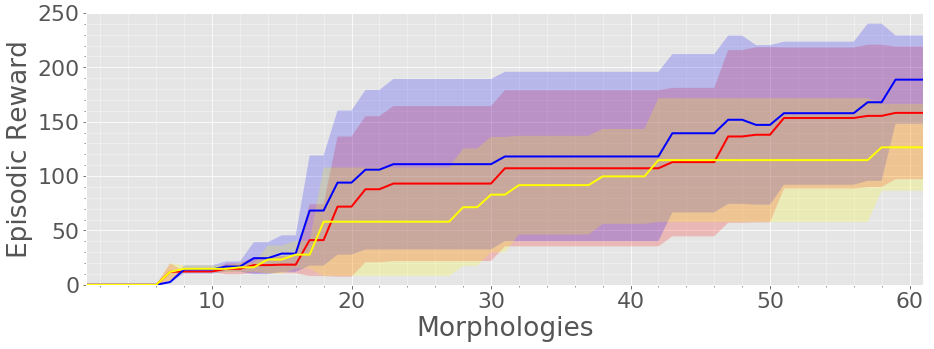}
        \caption{Morphology 4}
    \end{subfigure}
    
    \begin{subfigure}[t]{0.5\textwidth}
        \centering
        \includegraphics[width=\textwidth]{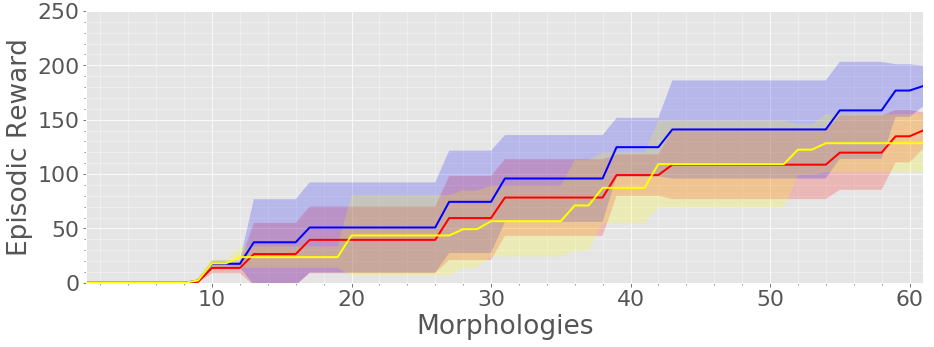}
        \caption{Morphology 5}
    \end{subfigure}%
    \caption{Performance of SG-MOrph on the Multi-Ped environment. Graphs show the average and standard deviation of the best performance seen in 100 episodes for each morphology-design combination, computed from five experiments. The first 10 morphologies/designs belong to the fixed initial design pool (Fig. \ref{Appendix::Fig::multiped::Initial}), thereafter 25 design optimization and 25 random design selections are performed. Figures a-e show the performance of SG-MOrph split into the five different morphological structures considered.}
    \label{Appendix::Fig::multiped::morphdata}
\end{figure*}

\begin{figure*}[h!]
    \centering
    \begin{subfigure}[t]{0.25\textwidth}
        \centering
        \includegraphics[width=\textwidth]{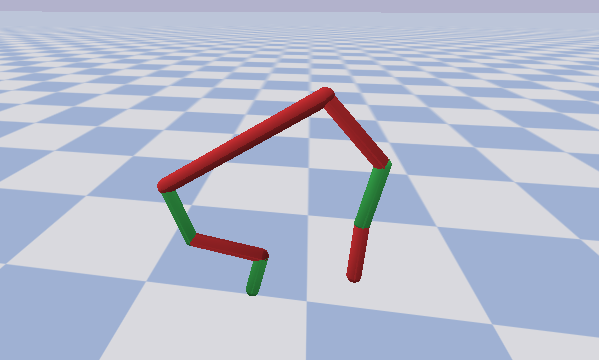}
        \caption{Morph. 1, Design 1}
    \end{subfigure}%
    ~ 
    \begin{subfigure}[t]{0.25\textwidth}
        \centering
        \includegraphics[width=\textwidth]{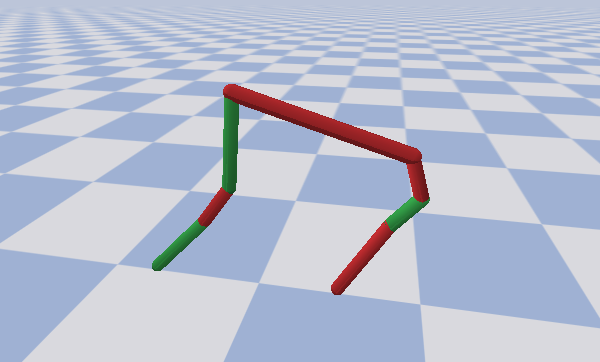}
        \caption{Morph. 1, Design 2}
    \end{subfigure}%
    ~
    \begin{subfigure}[t]{0.25\textwidth}
        \centering
        \includegraphics[width=\textwidth]{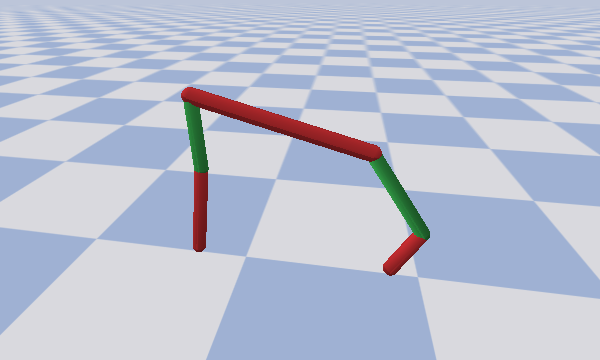}
        \caption{Morph. 2, Design 1}
    \end{subfigure}%
    ~ 
    \begin{subfigure}[t]{0.25\textwidth}
        \centering
        \includegraphics[width=\textwidth]{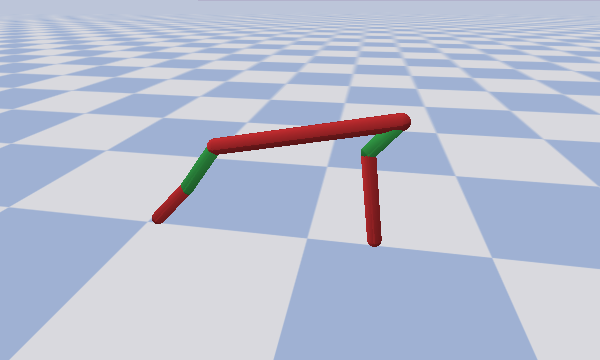}
        \caption{Morph. 2, Design 2}
    \end{subfigure}
    
    \begin{subfigure}[t]{0.25\textwidth}
        \centering
        \includegraphics[width=\textwidth]{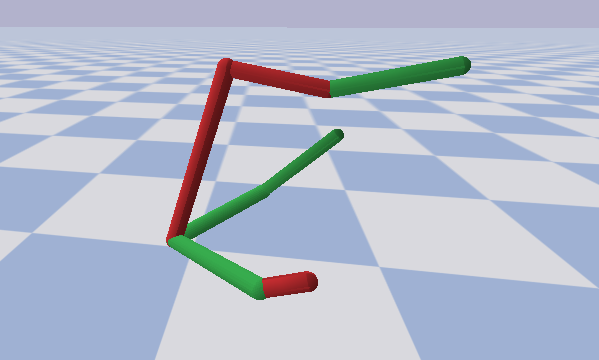}
        \caption{Morph. 3, Design 1}
    \end{subfigure}%
    ~ 
    \begin{subfigure}[t]{0.25\textwidth}
        \centering
        \includegraphics[width=\textwidth]{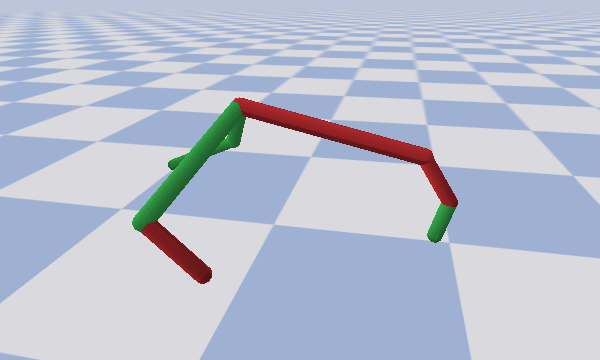}
        \caption{Morph. 3, Design 2}
    \end{subfigure}%
    ~
    \begin{subfigure}[t]{0.25\textwidth}
        \centering
        \includegraphics[width=\textwidth]{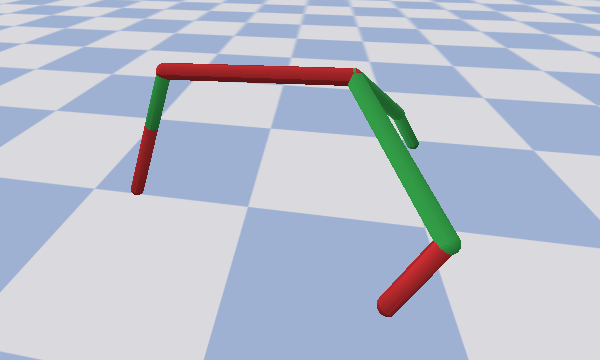}
        \caption{Morph. 4, Design 1}
    \end{subfigure}%
    ~ 
    \begin{subfigure}[t]{0.25\textwidth}
        \centering
        \includegraphics[width=\textwidth]{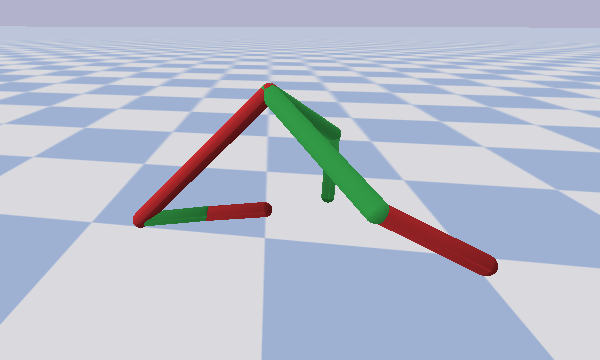}
        \caption{Morph. 4, Design 2}
    \end{subfigure}
    
    \begin{subfigure}[t]{0.25\textwidth}
        \centering
        \includegraphics[width=\textwidth]{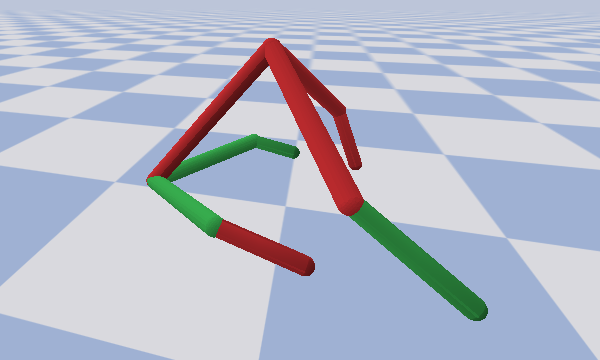}
        \caption{Morph. 5, Design 1}
    \end{subfigure}%
    ~ 
    \begin{subfigure}[t]{0.25\textwidth}
        \centering
        \includegraphics[width=\textwidth]{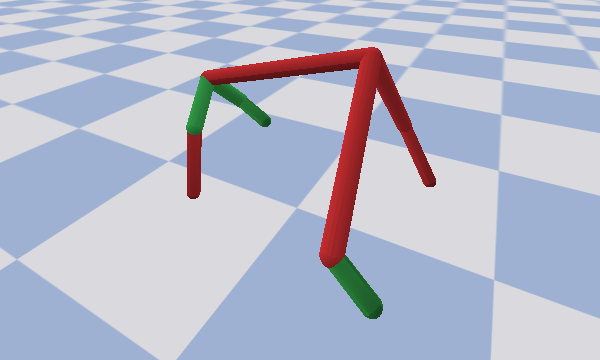}
        \caption{Morph. 5, Design 2}
    \end{subfigure}
    \caption{Initial designs pool for Multi-Ped consisting of five different morphologies, \ie graph structures, ranging from four to six degrees-of-freedom.}
    \label{Appendix::Fig::multiped::Initial}
\end{figure*}

The initial pool of morphology-design combinations was here 10, with five distinct morphological structures (Fig. \ref{Appendix::Fig::multiped::Initial}). 
If we split the performance of SG-MOrph by morphologies (see Fig.\ \ref{Appendix::Fig::multiped::morphdata}) we can see that this is the first environment in which the GNN policies are able to catch up, and even perform slightly better at the end of the training process in morphologies 3-5. 
Similarly, we can see that in this environment the random design selection baseline performs much better than in prior environments, although SG-MOrph is still able to outperform it. 
A possible explanation is that in this specific environment the number of designs which are delivering optimal performance for a morphology could be numerous enough, that they can be discoerved by random selection. 
Thus, we decided to repeat this experiment on an harder instance of this environment, which will be presented in the next section. 
Three designs which were found with a high fitness by SG-MOrph are presented in Figure  \ref{Appendix::Fig::multiped::optimal}. 

\clearpage
\subsection{Environment 4: Multi-Ped+Stairs}
\begin{figure*}[]
    \centering
    \begin{subfigure}[t]{0.2\textwidth}
        \centering
        \includegraphics[width=\textwidth]{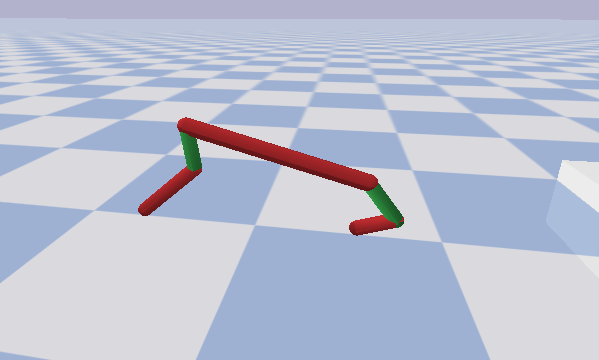}
        \caption{4 DoF}
    \end{subfigure}%
    ~ 
    \begin{subfigure}[t]{0.2\textwidth}
        \centering
        \includegraphics[width=\textwidth]{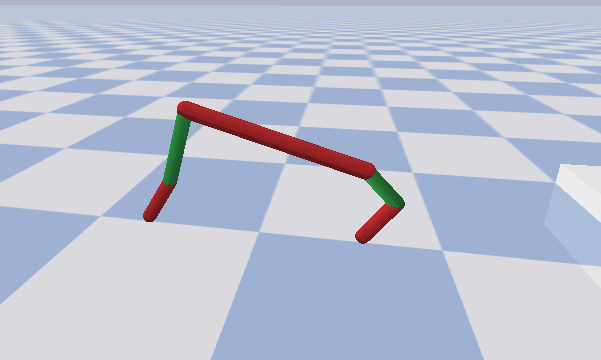}
        \caption{5 DoF}
    \end{subfigure}%
    ~
    \begin{subfigure}[t]{0.2\textwidth}
        \centering
        \includegraphics[width=\textwidth]{fig2/designs/best/stairs/stairs_2_cut.png}
        \caption{5 DoF}
    \end{subfigure}%
     \caption{Selection of optimized  morphologies found for Multi-Ped+Stairs.}
    \label{Appendix::Fig::stairs::optimal}
\end{figure*}

\begin{figure}[t!]
        \centering
        \includegraphics[width=0.5\textwidth]{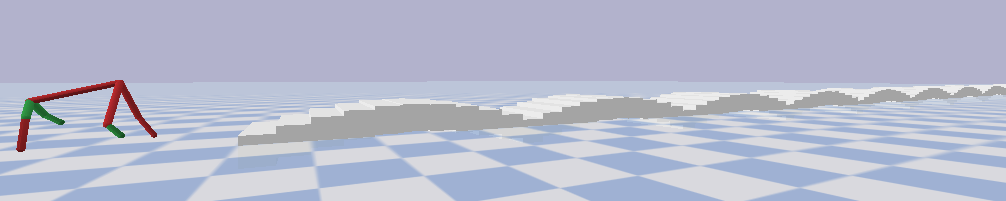}
        \caption{Simulation environment for Multi-Ped+Stairs.}
        \label{Appendix::Fig::stairs::env}
\end{figure}

\begin{figure*}[h!]
    \centering
    \begin{subfigure}[t]{0.5\textwidth}
        \centering
        \includegraphics[width=\textwidth]{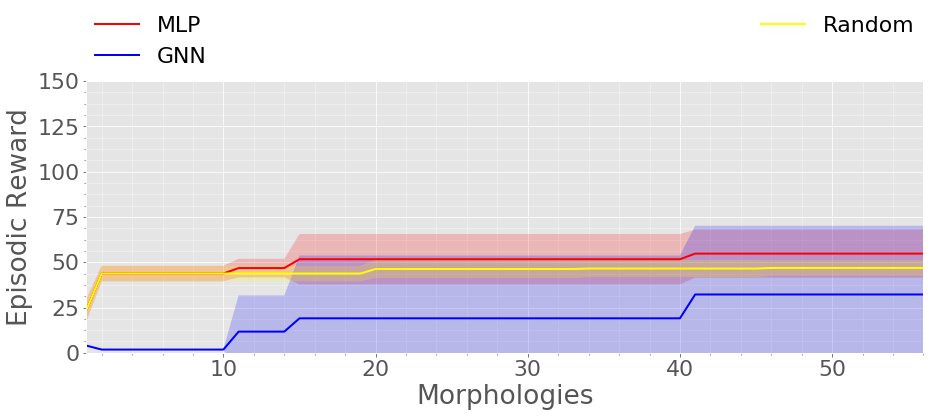}
        \caption{Morphology 1}
    \end{subfigure}%
    ~ 
    \begin{subfigure}[t]{0.5\textwidth}
        \centering
        \includegraphics[width=\textwidth]{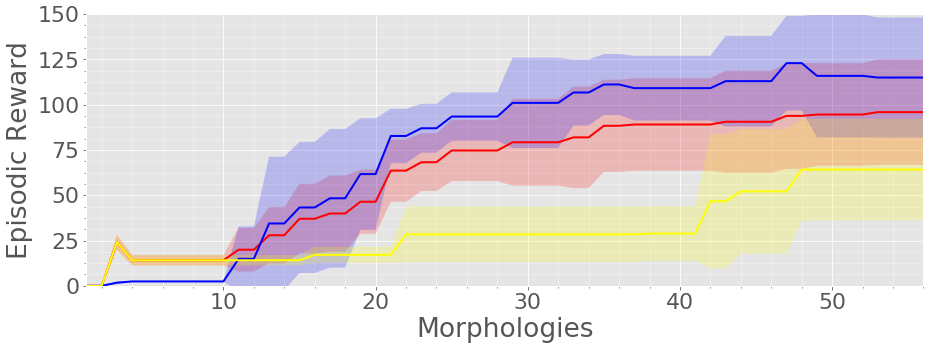}
        \caption{Morphology 2}
    \end{subfigure}
    
    \begin{subfigure}[t]{0.5\textwidth}
        \centering
        \includegraphics[width=\textwidth]{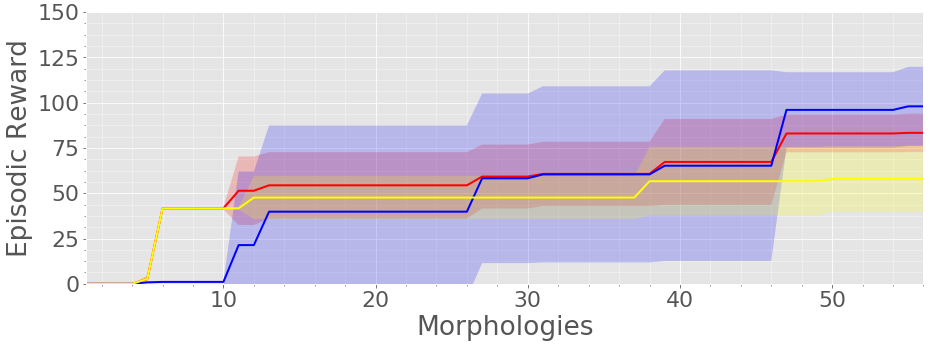}
        \caption{Morphology 3}
    \end{subfigure}%
    ~ 
    \begin{subfigure}[t]{0.5\textwidth}
        \centering
        \includegraphics[width=\textwidth]{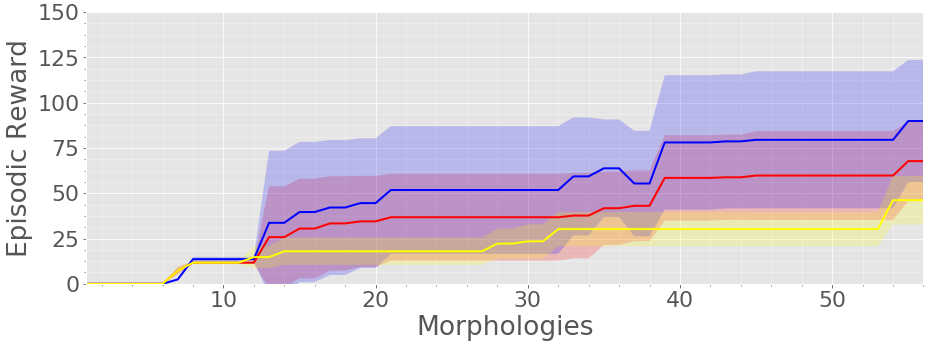}
        \caption{Morphology 4}
    \end{subfigure}
    
    \begin{subfigure}[t]{0.5\textwidth}
        \centering
        \includegraphics[width=\textwidth]{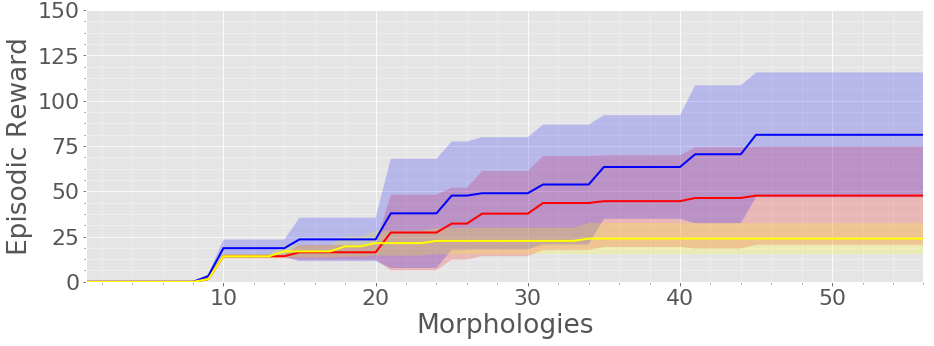}
        \caption{Morphology 5}
    \end{subfigure}%
    \caption{Performance of SG-MOrph on the Multi-Ped+Stairs environment. Graphs show the average and standard deviation of the best performance seen in 100 episodes for each morphology-design combination, computed from five experiments. The first 10 morphologies/designs belong to the fixed initial design pool (Fig. \ref{Appendix::Fig::multiped::Initial}), thereafter 25 design optimization and 25 random design selections are performed. Figures a-e show the performance of SG-MOrph split into the five different morphological structures considered.}
    \label{Appendix::Fig::stairs::all}
\end{figure*}

This environment is based on the previously introduced Multi-Ped environment, using the same morphological structures for optimization as well as reward functions and state spaces. 
Unlike the Multi-Ped environment, which had the agent traverse a flat terrain, the task in this environment is to navigate an uneven terrain with staircases as shown in Figure \ref{Appendix::Fig::stairs::env}. 
The agent has to navigate this terrain in a blind manner and does not receive any additional information, such as the location or current height of the closest step. 
In fact, we can see in Figure \ref{Appendix::Fig::stairs::all} that the performance of the random design selection baseline is much worse than in the Multi-Ped task, indicating that the set of optimal design parameter is much smaller and unlikely to be found by uniform sampling strategies. 
However, we find that on most morphologies the gap between GNN policies and MLP policies, the first performing better, widens, indicating room for improvement of SG-MOrph. 
This will be discussed further in one of the following sections. 
Finally, Figure \ref{Appendix::Fig::stairs::optimal} shows a selection of designs found to be optimal by SG-MOrph.

\subsection{Visualizing the Performance of the Design Space}
\begin{figure}[h!]
    \centering
    \begin{subfigure}[t]{0.5\textwidth}
        \centering
        \includegraphics[width=\textwidth]{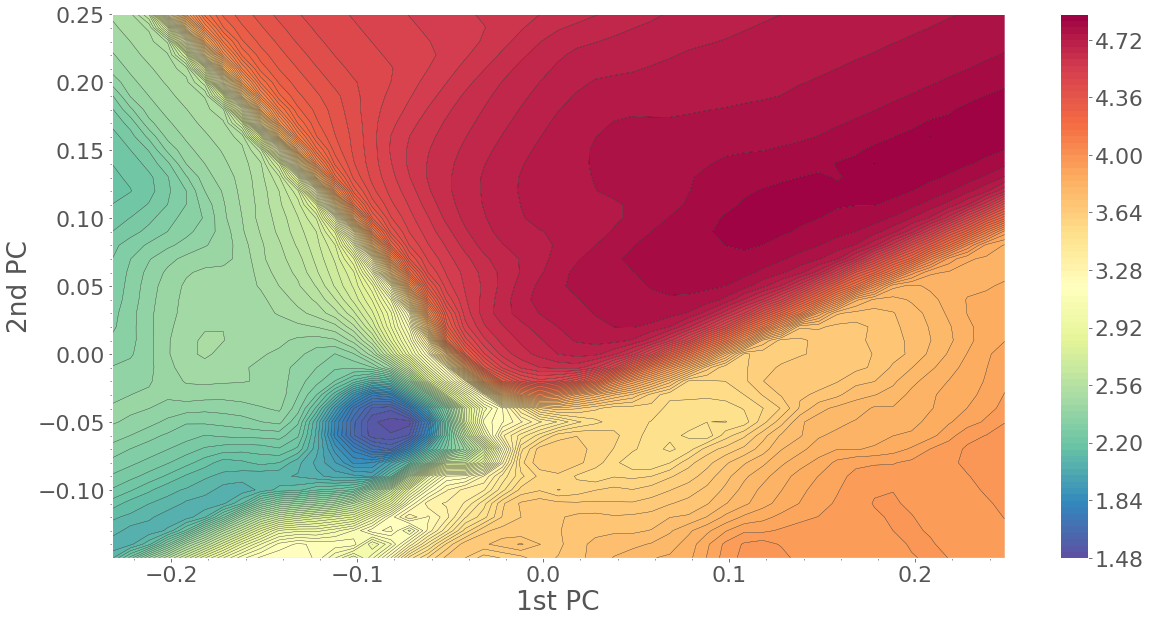}
        \caption{HalfCheetah Morph. 1}
    \end{subfigure}%
    \begin{subfigure}[t]{0.5\textwidth}
        \centering
        \includegraphics[width=\textwidth]{{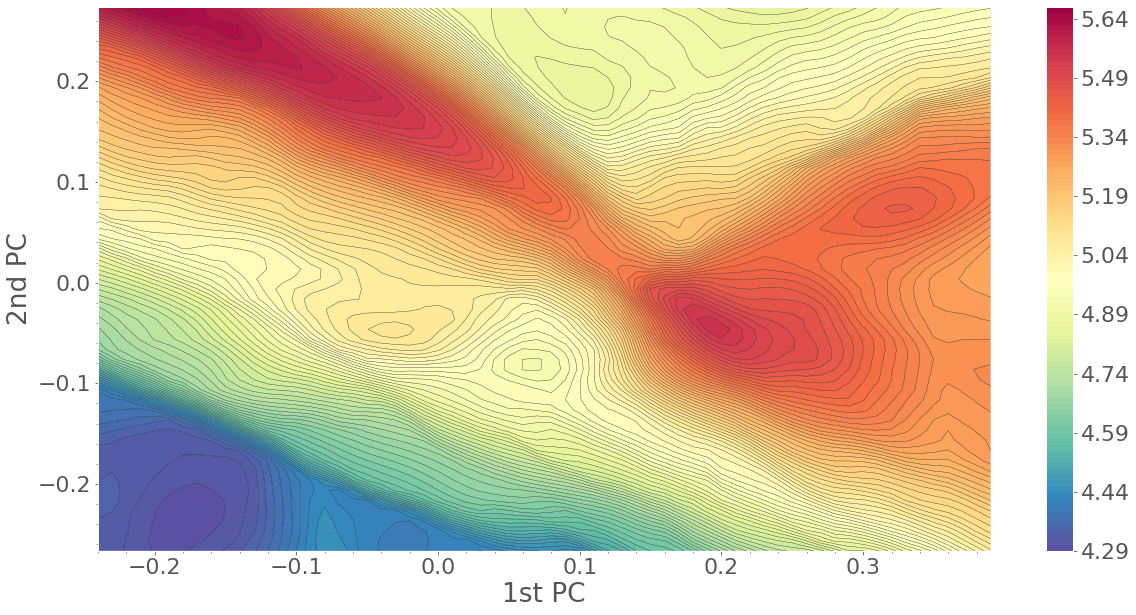}}
        \caption{HalfCheetah Morph. 5}
    \end{subfigure}%
    \caption{Visualization of the estimated performance of continuous designs variables for graph structures 1 and 5 in HalfCheetah after 64 morphology-design evaluations. The high-dimensional design parameter space was projected onto a two-dimensional latent space using Principal Component Analysis (PCA) on all designs evaluated on a specific morphology. The visualizations show only the valid range of design parameters. Estimated performance was computed by using the proposed objective function in Eq.\ \ref{Eq::mopt}.
    }
    \label{Appendix::Fig::HC::latents}
\end{figure}

One inherent property of the proposed framework is that the introduced objective function (Eq.\ \ref{Eq::mopt}) can be used to visualize the current believe of the graph neural networks for the performance of design variables in a given morphology/graph structure. 
Figure \ref{Appendix::Fig::HC::latents} shows as latent representation of the estimated performance of design variables for morphologies one and five in the HalfCheetah environment. 
We can see that the performance estimations for design variables are not uniform but the proposed objective funciton gives rise to complex structures and predictions in the design space. 
While this visualization property is not exploited in this paper, it opens the door for better integration of human designers in the automatic design adaptation process. 

\subsection{Limitations, Dirty Laundry and Future Work}
We see the main contribution of this paper in providing a first stepping stone for an effective framework combining classic MLP networks with graph neural networks for the fast and data-efficient co-adaptation of robotic designs and behaviour, suitable for direct real-world application. 
Nevertheless, based on the experiment conducted we were able to identify potential avenues for future improvements of the proposed algorithm. 
Experiments conducted on the two Multi-Ped environments show that while the proposed transfer-learning approach is able to jump-start MLP networks on a new design, it can be outperformed by GNN policies on certain environments. 
Our hypothesis is that the GNN actor is able to benefit in these harder environments from receiving training data seen from earlier successful prototypes, while the MLP actor is only being trained on data collected from the current prototype after the transfer-learning phase. 
A potential avenue to increase the performance of MLP policies further is to allow for the continuous flow of information and knowledge from GNN to MLP networks. 
This could be achieved, for example, by using loss functions combining both MLP and GNN critics, or by placing a prior on the MLP policies forcing them to stay in the vicinity of the GNN policy in the parameter or action space. 
Such approaches would still allow for the high-frequency evaluation of policies on robots, while using slower GNN networks only during the off-line training on an external computer. 
Furthermore, while we selected fixed pools of morphological structures similar to the experiments conducted in \cite{huang2020one}, of which we optimize the design parameters and thus having an infinite number of potential design prototypes, an obvious next step is allow for arbitrary graph structures. 
This would allow for the use of evolutionary algorithms mutating and adapting graph structures and could be performed without the need for extra simulations or real world experiments. 
Where this optional pool-adaptation could be performed is indicated in Algorithm 1 and Figure \ref{Appendix::Fig::Overview}. 

We think that the main obstacle for future developments in this problem area is the lack of available training environments which allow for the adaptation of discrete and continuous morphological and design parameters alike in an accessible and usable manner. 
Some prior work started to tackle this problem and provide open-source environments accessible to the research community. 
For example, Huang et al.\ introduce in \cite{huang2020one} a set of environments, each having a number of different agents with varying degrees-of-freedom, however without the ability to influence and change the physical parameters of agents. 
We aim to provide with this paper a first version of \textit{morphsim}, an simulation environment which allows for the easy adaptation of design parameters on a number of morphologies with an OpenAI-Gym-like interface and providing automatically generated graph objects usable in the Deep Graph Library. 
However, while the environment allows for an arbitrary but fixed number of morphological/graph structures, similarly to \cite{huang2020one}, it is currently not possible to auto-generate new morphological structures during execution. 
Future work on this environment will aim to not only generate graph representations from agents and allow the change of design parameters, but also to generate agents from graph structures - \ie the auto-generation of limbs and motors based on the nodes and edges present in a given graph. 
This in combination with tasks familiar to the reinforcement learning community, such as Walker, HalfCheetah and others, would allow for further research into the topic of co-adaptation and would allow other researchers to easily test and evaluate new algorithms and methods. 

\end{document}